\documentclass{article}

\usepackage{spconf,amsmath,amssymb,graphicx}

\usepackage[table]{xcolor}
\usepackage{multirow}
\usepackage{makecell}
\usepackage{booktabs} 

\usepackage[acronym,nonumberlist]{glossaries}

\usepackage{algorithmic}
\usepackage[ruled]{algorithm2e}

\usepackage[normalem]{ulem}

\usepackage{tikz}
\usepackage{pgfplots}
\pgfplotsset{width=10cm,compat=1.14}
\usepgfplotslibrary{statistics}
\usetikzlibrary{pgfplots.groupplots}

\usepackage{url}
\usepackage[hidelinks]{hyperref}
\hypersetup{
    colorlinks=true,
    linkcolor=black,
    filecolor=black,
    urlcolor=gray,
    citecolor=black,
    breaklinks=true
}

\newcommand*{\eg}{\textit{e.g.}}
\newcommand*{\ie}{\textit{i.e.}}

\newcommand{\mtrx}[1]{\mathbf{#1}}

\definecolor{Gray}{gray}{0.9}

\definecolor{nocs}{RGB}{0,0,0}
\definecolor{segdd}{RGB}{0, 153, 255}
\definecolor{iodenarrow}{RGB}{204, 51, 255}
\definecolor{iode}{RGB}{255, 0, 102}

\definecolor{iodeA}{RGB}{255,0,0}
\definecolor{iodeB}{RGB}{255,102,102}
\definecolor{iodeC}{RGB}{255,204,204}

\definecolor{iodenarrowA}{RGB}{0,255,0}
\definecolor{iodenarrowB}{RGB}{102,255,102}
\definecolor{iodenarrowC}{RGB}{204,255,204}

\definecolor{segddA}{RGB}{0,0,255}
\definecolor{segddB}{RGB}{102,102,255}
\definecolor{segddC}{RGB}{204,204,255}

\definecolor{nocsA}{RGB}{0,0,0}
\definecolor{nocsB}{RGB}{102,102,102}
\definecolor{nocsC}{RGB}{204,204,204}

\title{Multi-view shape estimation of transparent containers}
\name{Alessio Xompero$^{1}$
\thanks{This work is supported by the CHIST-ERA program through the project CORSMAL, under UK EPSRC grant EP/S031715/1 and Swiss NSF grant 20CH21{\_}180444.}
\thanks{Copyright 2020 IEEE. Published in the IEEE 2020 International Conference on Acoustics, Speech, and Signal Processing (ICASSP 2020), scheduled for 4-9 May, 2020, in Barcelona, Spain. Personal use of this material is permitted. However, permission to reprint/republish this material for advertising or promotional purposes or for creating new collective works for resale or redistribution to servers or lists, or to reuse any copyrighted component of this work in other works, must be obtained from the IEEE. Contact: Manager, Copyrights and Permissions / IEEE Service Center / 445 Hoes Lane / P.O. Box 1331 / Piscataway, NJ 08855-1331, USA. Telephone: + Intl. 908-562-3966.}, 
Ricardo Sanchez-Matilla$^{1}$,  Apostolos Modas$^{2}$, Pascal Frossard$^{2}$, Andrea Cavallaro$^{1}$}
\address{$^{1}$Centre for Intelligent Sensing, Queen Mary University of London, UK\\
  $^{2}$ LTS4, Ecole Polytechnique F\'ed\'erale de Lausanne (EPFL), Switzerland}
\begin{document}

\ninept
\newacronym{iode}{LoDE}{Localisation and Object Dimensions Estimator}
\newacronym{nocs}{NOCS}{Normalized Object Coordinate Space}
\newacronym{stocs}{StoCS}{Stochastic Congruent Sets}
\newacronym{dnn}{DNN}{Deep Neural Network}
\newacronym{pvnet}{PVNet}{Pixel-wise Voting Network}
\newacronym{lodeir}{LoDE-IR}{Localisation and Object Dimensions Estimator with Infrared}

\maketitle
\thispagestyle{empty}
\pagestyle{empty}

\begin{abstract} 
The 3D localisation of an object and the estimation of its properties, such as shape and dimensions, are challenging under varying degrees of transparency and lighting conditions. 
In this paper, we propose a method for jointly localising container-like objects and estimating their dimensions using two wide-baseline, calibrated RGB cameras. Under the assumption of circular symmetry along the vertical axis, we estimate the dimensions of an object with a generative 3D sampling model of sparse circumferences, iterative shape fitting and image re-projection to verify the sampling hypotheses in each camera using semantic segmentation masks.
We evaluate the proposed method on a novel dataset of objects with different degrees of transparency and captured under different backgrounds and illumination conditions. Our method, which is based on RGB images only, outperforms in terms of localisation success and dimension estimation accuracy a deep-learning based approach that uses depth maps.
\end{abstract}
\begin{keywords}
Object localisation, Dimension estimation, Transparency.
\end{keywords}
%

\section{Introduction}
\label{sec:intro}

Localising objects in 3D and estimating their properties (\eg~dimensions, shape), as well as their pose (location, orientation), is important for several robotic tasks, such as grasping~\cite{Wang2019IJARS,Saxena2008}, manipulation~\cite{Calli2017IJRR_YCB} and human-to-robot handovers~\cite{Medina2016}. However, everyday objects can widely vary in shape, size, material, and transparency, thus making the vision-based estimation of their properties a challenging problem. 
 
Existing methods for localising objects in 3D or estimating their 6 Degrees of Freedom (DoF)
pose rely on databases of 3D object models (\eg~CAD)~\cite{Wang2019CVPR_DenseFusion,Peng2019CVPR_PVNet,Mitash2018BMVC_StoCS,Qi2018CVPR_FrustumNet} or motion capture systems~\cite{Medina2016,Hartley2003,Kim2014}. To avoid using markers for motion capture, feature points~\cite{Lowe2004IJCV,Dusmanu2019CVPR_D2Net} can be localised in an image and matched against a 3D object model to estimate the object pose by solving a Perspective-n-Point (PnP) problem~\cite{Lepetit2009IJCV}. However, this strategy may fail when objects exhibit limited texture or are captured under unfavourable lighting conditions~\cite{Mitash2018BMVC_StoCS}. Approaches based on \acrfull{dnn} learn to estimate the 6~DoF object pose quite accurately, but their training requires large amount of data, usually annotated only for the high-level object class~\cite{Lin2018ECCV}, including depth information and/or \emph{known} dense 3D models in addition to colour images~\cite{Wang2019CVPR_DenseFusion,Peng2019CVPR_PVNet,Mitash2018BMVC_StoCS,Wang2019CVPR_NOCS,Yinlin2019}. For example, PoseCNN~\cite{Xiang2018RSS_PoseCNN}, DenseFusion~\cite{Wang2019CVPR_DenseFusion}, SegOPE~\cite{Yinlin2019} and PVNet~\cite{Peng2019CVPR_PVNet} are evaluated only with objects whose high-quality 3D models and depth were available~\cite{Xiang2018RSS_PoseCNN}, discarding testing objects that are transparent as the segmentation may fail or be inaccurate.
DenseFusion~\cite{Wang2019CVPR_DenseFusion} combines features obtained from RGB-D data to handle occlusions and inaccurate segmentation.
\acrfull{pvnet}~\cite{Peng2019CVPR_PVNet} estimates the pose of occluded or truncated objects with an uncertainty-driven PnP, learning a vector-field representation to localise a sparse set of 2D keypoints and their spatial uncertainty.
\acrfull{nocs}~\cite{Wang2019CVPR_NOCS} formulates this coordinate space to jointly estimates the 6~DoF pose and the dimensions (in the form of a 3D bounding box) of an object not seen during the training of the DNN (\eg~intra-class variability for object shape, size, and appearance). 
As most of these works target object pose estimation, related comprehensive reviews can be found in~\cite{Wang2019CVPR_DenseFusion,Peng2019CVPR_PVNet,Wang2019CVPR_NOCS,Yinlin2019}.
To handle textureless, translucent or reflective objects (\eg~wine glasses), whereas 3D reconstruction may perform poorly, Learning the Grasping Point (LGP)~\cite{Saxena2008} uses supervised training on synthetic images with annotated grasping regions and learns to identify in two or more images a few points that are good for grasping unknown objects in 3D. 
Table~\ref{tab:soa} summarises relevant works based on their assumptions; their targeted tasks, especially the object dimensions estimation in addition to the localisation in 3D; and their capability to handle transparent objects. 
Nevertheless, estimating the dimensions of these objects in 3D is still challenging. 

\begin{table}[t!]
    \centering
    \footnotesize
    \setlength\tabcolsep{2.6pt}
    \caption{Comparison of markerless methods for object localisation and dimensions estimation in 3D. 
    KEY -- Ref.:~reference; n3D:~no 3D object model; nD:~no depth; HLC:~known high-level object class; Loc.:~object localisation in 3D; Dim.:~object dimensions estimation in 3D; 3DM:~dimensions given by the 3D model.}
    \vspace{0.1cm}
    \begin{tabular}{rlcccccc}
    \specialrule{1.2pt}{0.2pt}{1pt}
    \textbf{Ref.} & \textbf{Method}  & \multicolumn{3}{c}{\textbf{Assumptions}} & \multicolumn{2}{c}{\textbf{Tasks}} & \multicolumn{1}{c}{\textbf{Transparency}}\\
    \cmidrule(lr){3-5}\cmidrule(lr){6-7}
     &   & n3D & nD & HLC  & Loc. & Dim. &  \\
    \specialrule{1.2pt}{0.2pt}{1pt}
    \cite{Saxena2008}     & LGP & \checkmark & \checkmark &  & \checkmark & & \checkmark \\
    \cite{Li2018ECCV}     & DeepIM &  & \checkmark & & \checkmark & 3DM & \\
    \cite{Mitash2018BMVC_StoCS} & \acrshort{stocs} & & &  \checkmark & \checkmark & 3DM & \\
    \cite{Peng2019CVPR_PVNet}       & \acrshort{pvnet}  & & \checkmark & \checkmark & \checkmark & 3DM &  \\
    \cite{Wang2019CVPR_DenseFusion} & DenseFusion & & &  \checkmark & \checkmark & 3DM &  \\
    \cite{Yinlin2019}     & SegOPE & & \checkmark &  \checkmark & \checkmark & 3DM &  \\
    \cite{Wang2019CVPR_NOCS} & \acrshort{nocs}  & \checkmark & & \checkmark & \checkmark & \checkmark & \\
    \midrule
    & \textbf{LoDE} & \checkmark & \checkmark  & \checkmark & \checkmark & \checkmark & \checkmark \\
     \specialrule{1.2pt}{0.2pt}{1pt}
    \end{tabular}
    \label{tab:soa}
    \vspace{-15pt}
\end{table}

In this paper, we propose LoDE (Localisation and object Dimensions Estimator)\footnote{\scriptsize{\url{http://corsmal.eecs.qmul.ac.uk/LoDE.html}}}, a method that estimates the dimensions of container-like objects, such as cups, drinking glasses and bottles, using two calibrated RGB cameras, whose poses are known.
LoDE localises the 3D centroid of the object from 2D centroids estimated from semantic segmentation masks. As most of these containers have a circular symmetry along their vertical axis, LoDE hypothesises an initial model with a set of circumferences sampled around the 3D centroid at different heights. Then, the model iteratively fits to the object by reducing the radius for sampling the circumferences until each circumference is verified within the object mask in each camera. We also collected a novel dataset with objects of different shapes and degrees of transparency, under varying lighting conditions and backgrounds.

\section{Localisation and dimension estimation}
\label{sec:method}

We propose a generative 3D sampling model to estimate the shape of an object and, as by-product, its dimensions, assuming the object to be circular symmetric with respect to its vertical axis.
We represent the object as $\mtrx{O} = (x,y,z,w,h) \in \mathbb{R}^5$, where ${\mtrx X} = (x,y,z) \in \mathbb{R}^3$ is the location of its centroid in 3D, and $h$ and $w$ are the height and the largest width, respectively. 
Let $I^c$ represent the camera views, where the object is observed, and $\mtrx{C}^c$  be the 3D pose of each camera whose calibration is modelled by the intrinsic parameters $\boldsymbol{\theta}^c$, consisting of focal length and principal point, with $c\in\{1,2\}$. 

As the object location and shape in 3D are unknown, we propose an iterative multi-view 3D-2D shape fitting via  projective geometry~\cite{Hartley2003} (see Fig.~\ref{fig:abst}). The object is first detected in each image $I^c$ via semantic segmentation:
\begin{figure}[t!]
    \centering
    \includegraphics[width=0.9\columnwidth]{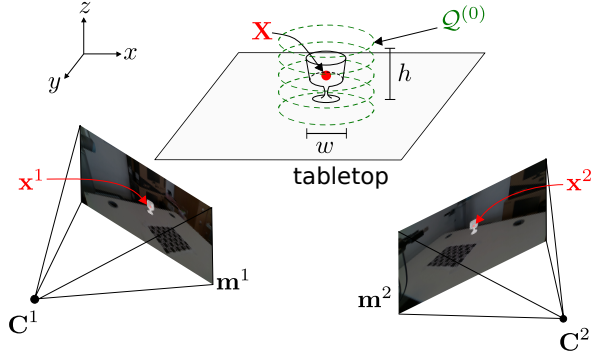}
    \vspace{-0.2cm}
    \caption{Two cameras capture an object from different viewpoints. Given only RGB images and camera poses, we estimate the width $w$ and height $h$ of the object without relying on 3D object models, depth information, or markers. 
    The proposed method, LoDE, localises the object centroid in 3D, $\mtrx{X}$, from the 2D centroids, $\mtrx{x}^1$ and $\mtrx{x}^2$, estimated on the segmented images, $\mtrx{m}^1$ and $\mtrx{m}^2$, and then samples a set of sparse 3D points, $\mathcal{Q}^{(0)}$, belonging to circumferences centred at the centroid location and at different heights, to fit the object shape with an iterative 3D-2D algorithm. 
    }
    \label{fig:abst}
    \vspace{-10pt}
\end{figure}
%
\begin{equation}
        D: \{0,\dots,255\}^{W,H,C} \rightarrow  \{0,1\}^{W,H},
\end{equation}
where $W$, $H$, $C$ are the image width, height and number of colour channels, respectively, and  $\mtrx{m}^c = D(I^c) \in \{0,1\}^{W,H}$ a binary feature map representing the segmented object.

Finding pixel correspondences between views when objects are textureless is challenging and ambiguities can lead to inaccurate estimations of the object location in 3D and its dimensions. We instead estimate the 2D centroid ${\mtrx{x}^c}$ of the segmented object with the intensity centroid method~\cite{Rosin1999CVIU} through the definition of the moments within a local image area. Then, we triangulate the two 2D centroids to estimate the object centroid in 3D~\cite{Hartley2003}:
\begin{equation}
        \tilde{\mtrx{X}} = \tau(\mtrx{x}^1, \mtrx{x}^2, \mtrx{C}^1, \mtrx{C}^2, \boldsymbol{\theta}^1, \boldsymbol{\theta}^2),
\end{equation}
where $\tau$ is the triangulation operator.

To estimate the object shape, we initialise around its estimated 3D centroid a cylindrical model that iteratively fits the object shape as observed by the cameras. For each iteration $i$, our approach samples $L$ circumferences of radius $r^{(i)}$, centred at the estimated object 3D location $\tilde{\mtrx{X}}$ and with varying height $z_l$, $l=1,\ldots, L$, 
\begin{equation}
       \mathcal{C}^{(i)} = \{(r_l^{i}, z_l, \nu_l)\}_{l=1:L},
\end{equation}
where $\nu_l \in \{0,1\}$ indicates whether a circumference lies within the object mask of both cameras. For each circumference $l$, we sample a set of $N$ sparse 3D points, 
\begin{equation}
    \mathcal{Q}^{(i)}_l =  \{\mtrx{Q}^{(i)}_{n,l} = (x_{n,l}, y_{n,l},z_{l})\}_{n=1:N},
    \label{eq:setpoints}
\end{equation}
and the set of all sampled 3D points is $\mathcal{Q}^{(i)} = \{\mathcal{Q}_l^{(i)}\}_{l=1:L}$.
We project the sampled 3D points onto the image of both cameras as
\begin{equation}
    \mtrx{u}_{n,l}^c = \pi({\bf Q}_{n,l}^{(i)}, \mtrx{C}^c, \boldsymbol{\theta}^c),
    \label{eq:proj}
\end{equation} 
where $\pi(\cdot) : \mathbb{R}^3 \rightarrow \mathbb{R}^2$ is the projection function~\cite{Hartley2003}.
Then, we verify if all the points belonging to circumference $l$, $\mathcal{Q}^{(i)}_l$, lie within the object mask of both cameras,
\begin{equation}
    \eta = \sum_{n=1}^N {\bf m}^1 ({\bf u}_{n,l}^1) + {\bf m}^2 ({\bf u}_{n,l}^2),
\end{equation}
and if the condition is satisfied (\ie~$\eta = 2N$), we set the corresponding flag as converged, \ie~$\nu_l=1$.
\begin{figure}[t!]
    \centering
    \footnotesize
    \setlength{\tabcolsep}{4pt}
    \begin{tabular}{ccc}
        $i=0$ & $i=207$ & $i=295$ \\
        \includegraphics[height=0.21\columnwidth, trim={280px 60px 160px 120px},clip]{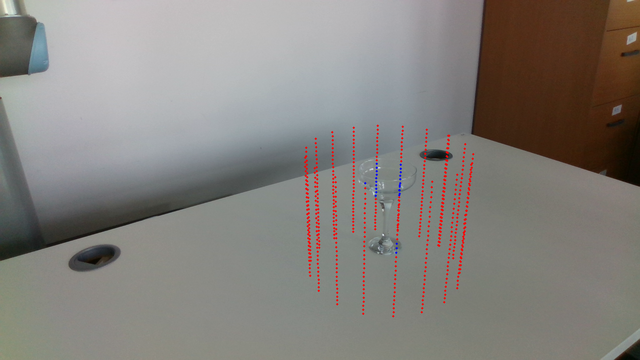}
        & 
        \includegraphics[height=0.21\columnwidth, trim={280px 60px 160px 120px},clip]{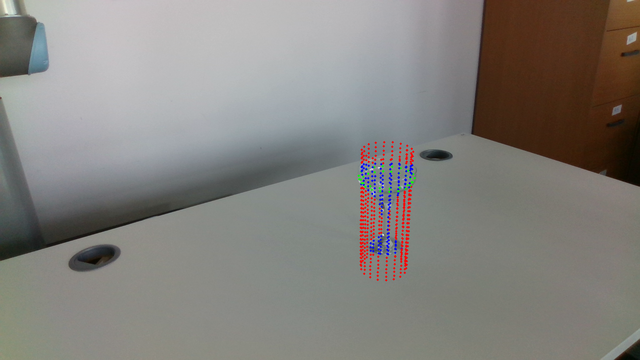}
        & 
        \includegraphics[height=0.21\columnwidth, trim={280px 60px 160px 120px},clip]{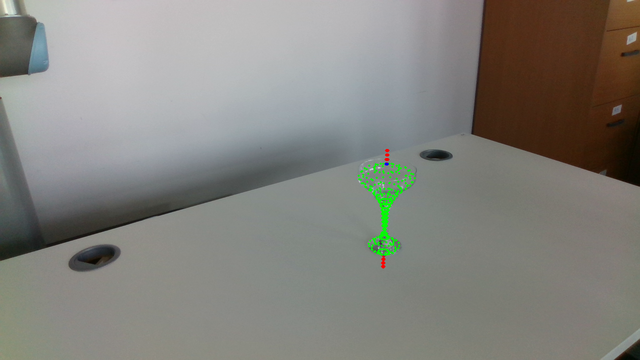}
        \\
        \includegraphics[height=0.21\columnwidth, trim={145px 53px 300px 135px},clip]{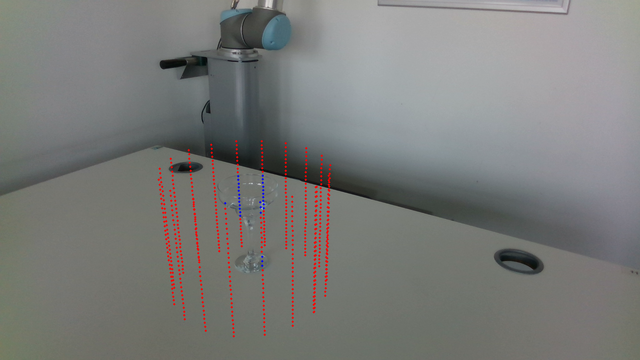}
        & 
        \includegraphics[height=0.21\columnwidth, trim={145px 53px 300px 135px},clip]{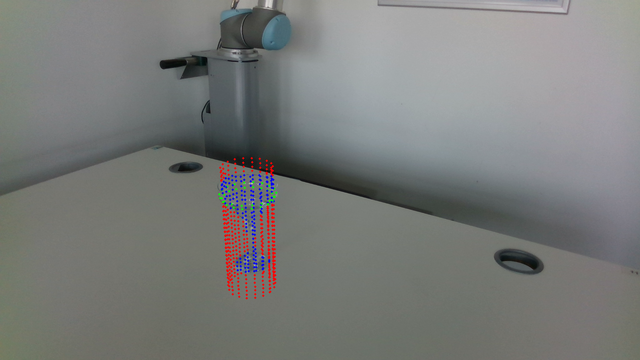}
        & 
        \includegraphics[height=0.21\columnwidth, trim={145px 53px 300px 135px},clip]{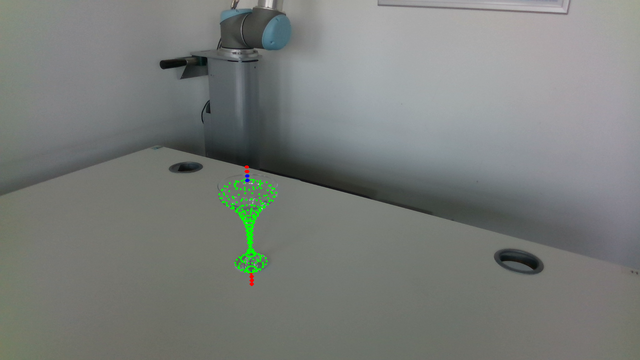} \\
        $r=150.00$~mm & $r=46.15$~mm & $r=2.00$~mm \\
    \end{tabular}
    \caption{Initialisation, sampled iteration and convergence of the 3D-2D shape fitting of a drinking glass (top: left camera, bottom: right camera).
    Legend: $i$ iteration number, $r$ radius of currently sampled circumference,
    \protect\tikz \protect\fill[red,fill=red] (1,1) circle (0.5ex);~projected points lying outside the segmentation mask,
    \protect\tikz \protect\fill[blue,fill=blue] (1,1) circle (0.5ex);~projected points lying inside the segmentation mask and
    \protect\tikz \protect\fill[green,fill=green] (1,1) circle (0.5ex);~projected points whose circumference fits the shape of the object (inside the segmentation mask of both cameras).
    }
    \label{fig:shapefitting}
    \vspace{-10pt}
\end{figure}
For iteration $i+1$, we decrease the radius $r_l^{(i+1)}$ and re-sample the 3D circumference points, $\mathcal{Q}^{(i+1)}$. Points belonging to circumference $l$ and with $\nu_l=1$ are not re-sampled. This iterative 3D-2D shape fitting terminates when either all $\nu_l = 1$ or $r_l^{i+1} < \rho$, where $\rho$ is the minimum radius that is sampled. Fig.~\ref{fig:shapefitting} shows as example three iterations of the shape fitting for a transparent drinking glass. 

Finally, to estimate the dimensions of the object, we select among the converged circumferences, $\mathcal{V} = \{ (r_l, z_l, \nu_l) | \nu_l = 1\} \subset \mathcal{C}$, the one with the largest radius $r^*$ and the ones with maximum and minimum heights, $z^*$ and $\bar{z}$, respectively. The estimated largest object width is $\tilde{w}=2r^*$ and the object height is $\tilde{h} = z^* - \bar{z}$.
\section{The CORSMAL Containers dataset}
\label{sec:dataset}

\begin{figure}[t!]
    \centering
    \setlength\tabcolsep{0pt}
    \begin{tabular}{cccccccc}
      \includegraphics[height=1.9cm]{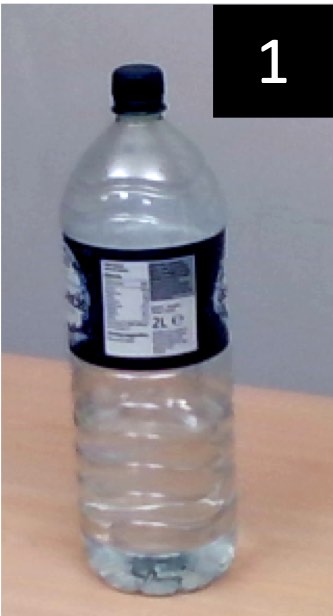} &
      \includegraphics[height=1.9cm]{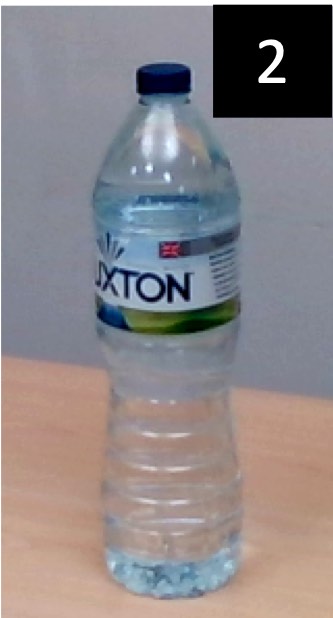} &
      \includegraphics[height=1.9cm]{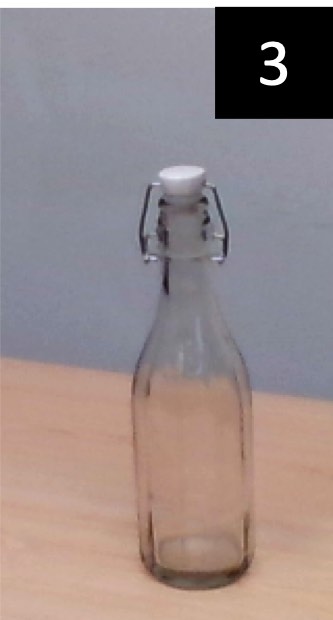} &
      \includegraphics[height=1.9cm]{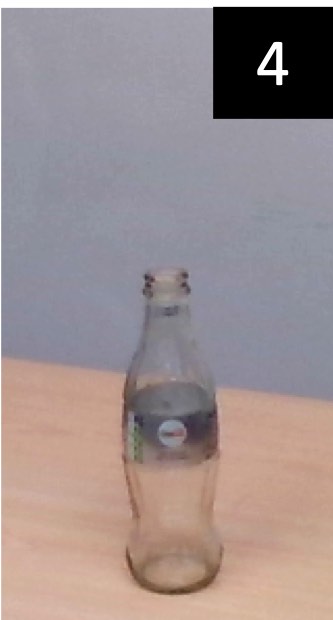} &
      \includegraphics[height=1.9cm]{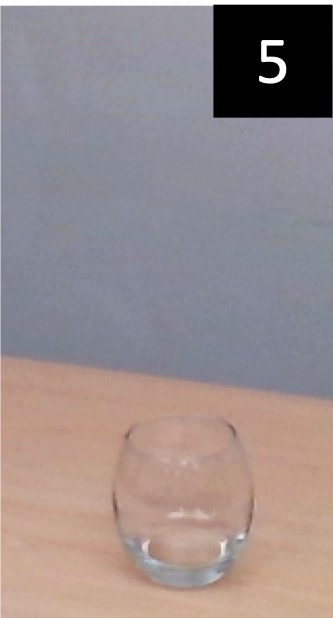} &
      \includegraphics[height=1.9cm]{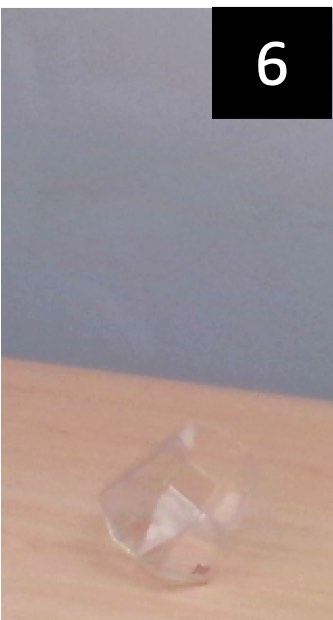} &
      \includegraphics[height=1.9cm]{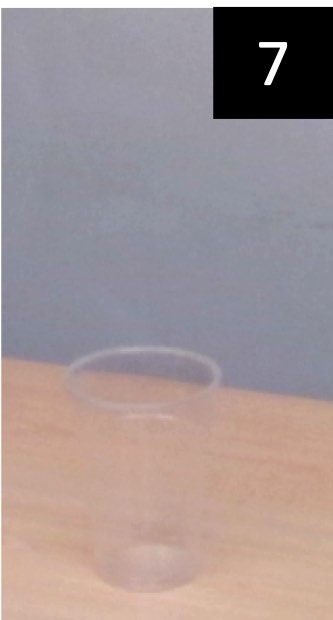} &
      \includegraphics[height=1.9cm]{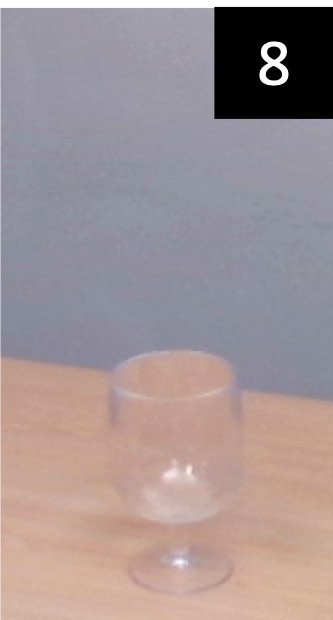} \\
      \includegraphics[height=1.9cm]{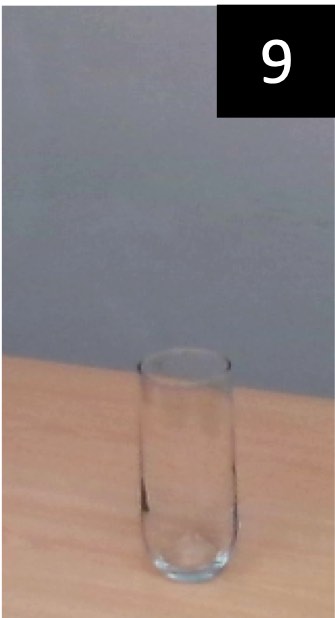} &
      \includegraphics[height=1.9cm]{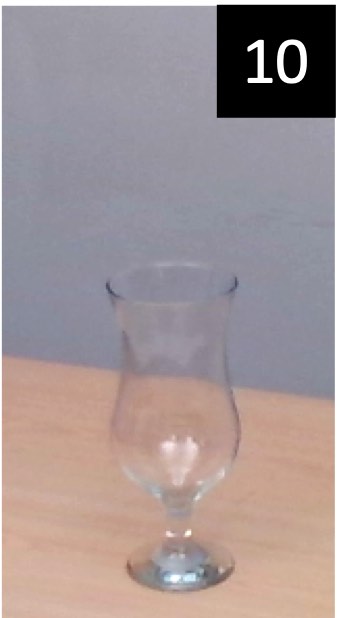} &
      \includegraphics[height=1.9cm]{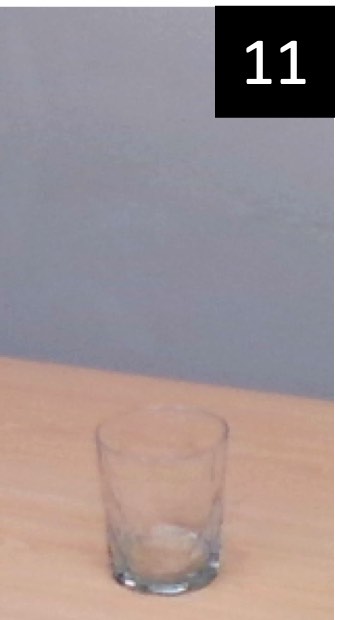} &
      \includegraphics[height=1.9cm]{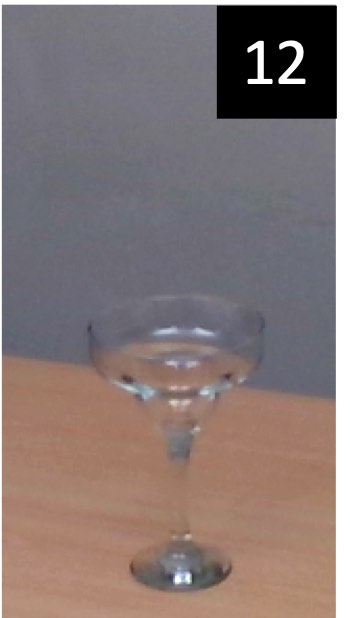} &
      \includegraphics[height=1.9cm]{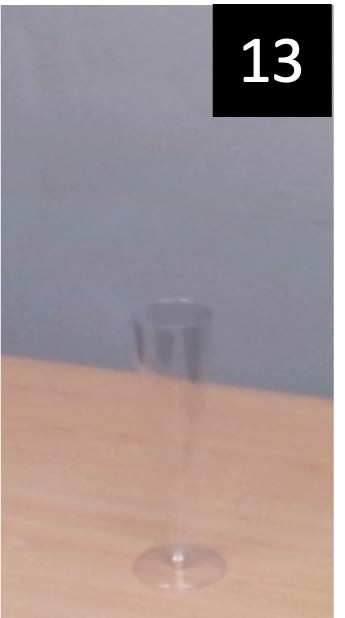} &
      \includegraphics[height=1.9cm]{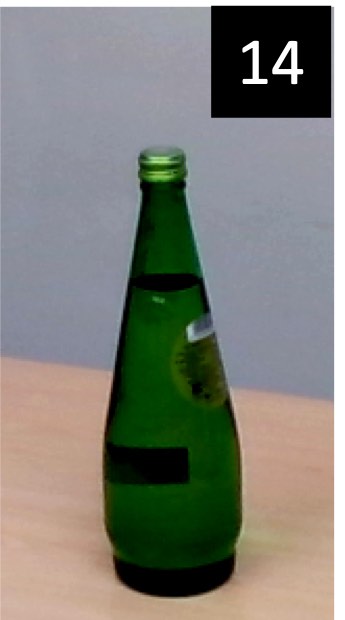} &
      \includegraphics[height=1.9cm]{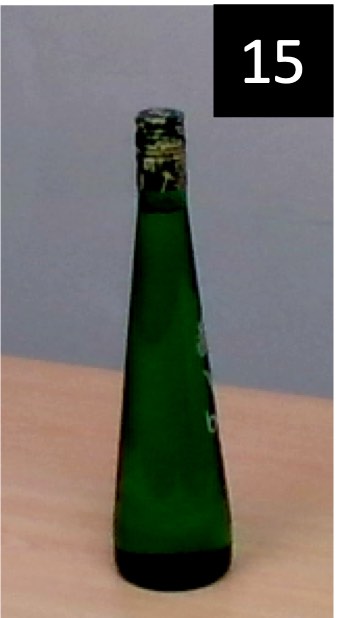} &
      \includegraphics[height=1.9cm]{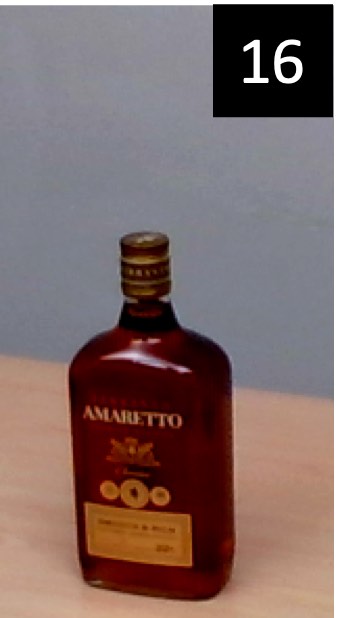} \\
      \includegraphics[height=1.9cm]{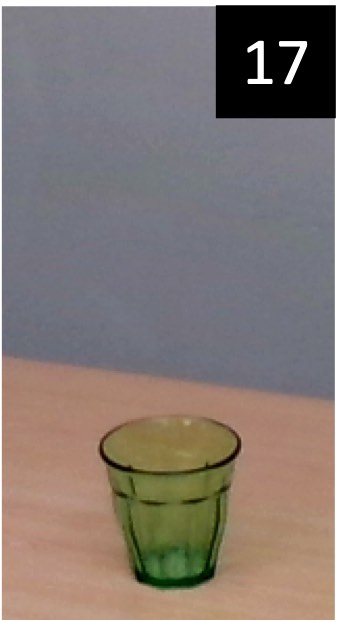} &
      \includegraphics[height=1.9cm]{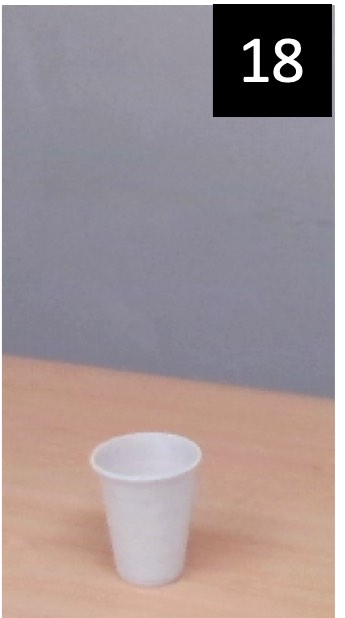} &
      \includegraphics[height=1.9cm]{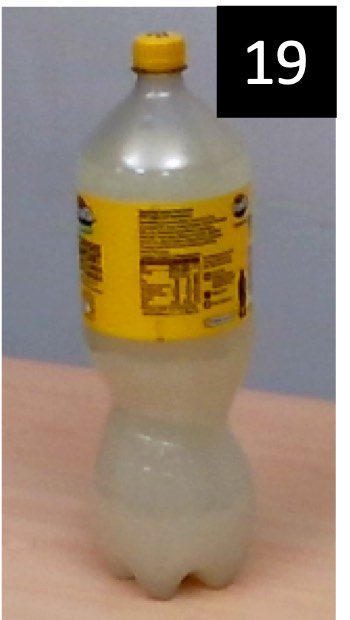} &
      \includegraphics[height=1.9cm]{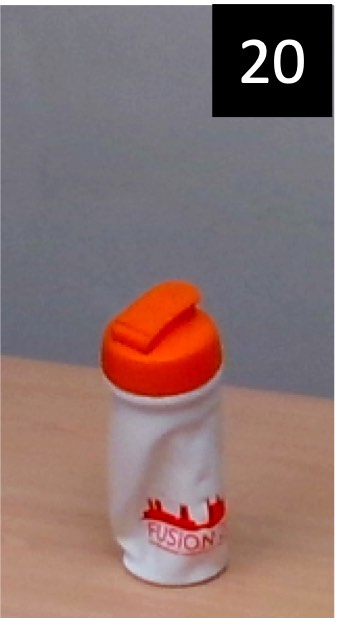} &
      \includegraphics[height=1.9cm]{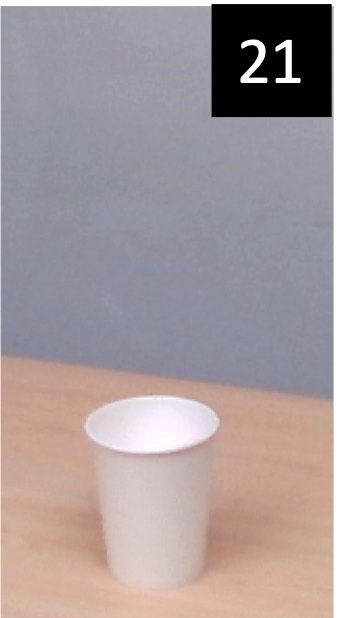} &
      \includegraphics[height=1.9cm]{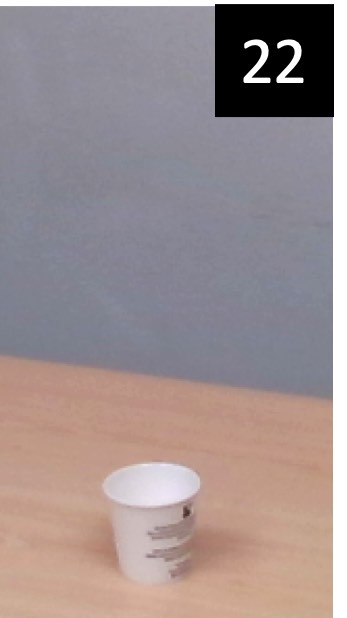} &
      \includegraphics[height=1.9cm]{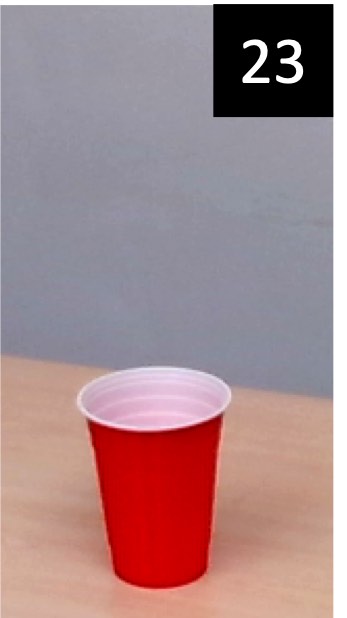} &
    \end{tabular}
    \vspace{-0.2cm}
    \caption{Objects in the CORSMAL Container dataset.
    Objects 1 to 13 (transparent); 14 to 18 (translucent); 19 to 23 (opaque). Note that crops are taken from images acquired with the same camera view.
    }
    \label{fig:dataset}
    \vspace{-10pt}
\end{figure}

We collect a set of images using 23 containers for liquids: 5 cups, 9 drinking glasses and 9 bottles (see Fig.~\ref{fig:dataset}). These objects are made of plastic, glass or paper, with different degrees of transparency and arbitrary shapes. The dataset contains 3 objects that do not have circular symmetry, \eg~object 6 (diamond-shaped glass), object 16 (amaretto bottle) and object 20 (deformed water-bottle).

We placed each object on a table and we acquired RGB, depth and stereo infrared (IR) images (1280$\times$720 pixels) with two Intel RealSense D435i cameras, located approximately at 40~cm from the object. RGB and depth images are spatially aligned. The cameras are calibrated and localised with respect to a calibration board.
We acquired the images in two rooms with different lighting and background conditions.
The first setup is an \emph{office} with natural light from a window  and objects placed on a table of size 160x80~cm and height 82~cm. The second setup is a \emph{studio}-like room with no windows, where we used either ceiling lights or artificial studio-like lights to illuminate a table of size 60x60~cm and height 82~cm.

To acquire multiple images of the same object under different backgrounds, we capture data with the tabletop uncovered and then covered with two different tablecloths. We collected in total 207 configurations that are combinations of objects (23), backgrounds (3) and lighting conditions (3), resulting in 414~RGB images, 414~depth images and 828~IR images. 
We annotated the largest width and height of each object with a digital caliper (0-150~mm, $\pm$0.01~mm) and a measuring tape (0-10~m, $\pm$0.001~m).

\section{Evaluation and results}
\label{sec:validation}

We compare LoDE with \acrshort{nocs}~\cite{Wang2019CVPR_NOCS}, a state-of-the-art \acrshort{dnn}-based approach that uses RGB-D data; and two baselines, which do not require 3D object models and can estimate object dimensions. One baseline uses segmentation on RGB-D data (SegDD). The other baseline is our approach applied to a stereo IR camera with narrow-baseline on a single device (\acrshort{lodeir}). SegDD partially replicates the initial part of several \acrshort{dnn}-based approaches~\cite{Wang2019CVPR_DenseFusion,Wang2019CVPR_NOCS,Xiang2018RSS_PoseCNN}, by using semantic segmentation and then back-projecting in 3D the pixels belonging to the object of interest, using the distance estimation of the depth image. The object dimensions are estimated from the most external points along the x-axis and y-axis, respectively (camera coordinate system). 
Note that while LoDE is multi-view, \acrshort{nocs}, SegDD and \acrshort{lodeir} are single-view. Thus, we report the results of single-view methods as the concatenation of the results from the two cameras.
Note that we do not compare with other approaches for 6~DoF pose estimation, (\eg~DenseFusion~\cite{Wang2019CVPR_DenseFusion}), or 3D Object Detection, (\eg~FrustumNet~\cite{Qi2018CVPR_FrustumNet}), as they require the exact 3D model of each object which is not the case of study of this work.

For the semantic segmentation, SegDD, \acrshort{lodeir} and LoDE adopt Mask R-CNN~\cite{He2017ICCV_MaskRCNN} trained on the MS COCO dataset~\cite{Lin2018ECCV} of which we consider the classes \textit{cup}, \textit{wine glass}, \textit{bottle} and \textit{vase}. For both \acrshort{lodeir} and LoDE, we set $L=500$ circumferences, separated by 1~mm on height and composed of $N=20$ points each (18$^\circ$ between point pairs) and we sample the radius of the circumferences, $r$, across iterations with the following schedule: $150.0, 149.5, \dots, 1.5, \rho$ (mm), with a minimum circumference radius of $\rho=1.0$~mm to fit the shape of objects that have a thin stem, (\eg~object 12, margarita glass, or object 8, plastic wine glass).

As performance measures, we compute the absolute error between the estimated and annotated width and height of the objects, and the Localisation Success Ratio (LSR), which measures the number of successful object localisations over the number of configurations (either the total number of configurations or a subset).

\pgfplotsset{
    every non boxed x axis/.style={},
    boxplot/every box/.style={solid,ultra thin,black},
    boxplot/every whisker/.style={solid,ultra thin,black},
    boxplot/every median/.style={solid,very thick, red},
}
\pgfplotstableread{IODE_errorH_transp0.txt}\IODEHopaque
\pgfplotstableread{IODE_errorH_transp1.txt}\IODEHtranslucid
\pgfplotstableread{IODE_errorH_transp2.txt}\IODEHtransparent
\pgfplotstableread{IODE_errorW_transp0.txt}\IODEWopaque
\pgfplotstableread{IODE_errorW_transp1.txt}\IODEWtranslucid
\pgfplotstableread{IODE_errorW_transp2.txt}\IODEWtransparent
\pgfplotstableread{IODE_errorH_transparency.txt}\IODEHtrans
\pgfplotstableread{IODE_errorW_transparency.txt}\IODEWtrans
\pgfplotstableread{IODE_narrow_errorH_transparency.txt}\IODEnarrowHtrans
\pgfplotstableread{IODE_narrow_errorW_transparency.txt}\IODEnarrowWtrans
\pgfplotstableread{SegDD_errorH_transparency.txt}\SegDDHtrans
\pgfplotstableread{SegDD_errorW_transparency.txt}\SegDDWtrans
\pgfplotstableread{NOCS_errorH_transparency.txt}\NOCSHtrans
\pgfplotstableread{NOCS_errorW_transparency.txt}\NOCSWtrans
\begin{figure}[t!]
    \centering
    \begin{tikzpicture}
      \begin{axis}[
        width=\columnwidth,
        xmin=0, xmax=24,
        y=0.1mm,
        xtick={1,2,3,4,5,6,7,8,9,10,11,12,13,14,15,16,17,18,19,20,21,22,23},
        xticklabels={},
        ymin=0,ymax=100,
        ytick={0,50,100},
        ylabel={LSR [\%]},
        label style={font=\footnotesize},
        tick label style={font=\footnotesize},
        ]
        \addplot[nocs, only marks, mark=square*, mark options={nocs,scale=0.5}] coordinates {
        (1,100.00)	(14,100.00)	(15,100.00)	(16,100.00)	(17,100.00)	(3,100.00)	(2,100.00)	(9,44.44)	(5,33.33)	(6,0.00)	(23,100.00)	(19,88.89)	(7,38.89)	(21,94.44)	(22,72.22)	(8,61.11)	(4,100.00)	(20,83.33)	(10,77.78)	(11,50.00)	(12,11.11)	(13,5.56)	(18,100.00)	
        };
        \addplot[segdd, only marks, mark=square*, mark options={segdd,scale=0.5}] coordinates {
        (1,100.00)	(14,100.00)	(15,100.00)	(16,100.00)	(17,100.00)	(3,100.00)	(2,100.00)	(9,94.44)	(5,61.11)	(6,0.00)	(23,100.00)	(19,100.00)	(7,55.56)	(21,100.00)	(22,100.00)	(8,88.89)	(4,100.00)	(20,100.00)	(10,100.00)	(11,100.00)	(12,100.00)	(13,72.22)	(18,100.00)
        };
        \addplot[iodenarrow, only marks, mark=square*, mark options={iodenarrow,scale=0.5}] coordinates {
        (1,72.22)	(14,100.00)	(15,83.33)	(16,16.67)	(17,100.00)	(3,100.00)	(2,83.33)	(9,16.67)	(5,11.11)	(6,0.00)	(23,66.67)	(19,72.22)	(7,27.78)	(21,83.33)	(22,27.78)	(8,33.33)	(4,94.44)	(20,72.22)	(10,66.67)	(11,61.11)	(12,66.67)	(13,0.00)	(18,100.00)	
        };
        \addplot[iode, only marks, mark=square*, mark options={iode,scale=0.5}] coordinates {
        (1,100.00)	(14,100.00)	(15,100.00)	(16,100.00)	(17,100.00)	(3,100.00)	(2,100.00)	(9,88.89)	(5,44.44)	(6,0.00)	(23,100.00)	(19,100.00)	(7,33.33)	(21,100.00)	(22,100.00)	(8,77.78)	(4,100.00)	(20,100.00)	(10,100.00)	(11,100.00)	(12,100.00)	(13,55.56)	(18,100.00)
        };
      \end{axis}
    \end{tikzpicture}
       \begin{tikzpicture}
        \begin{axis}[
            axis x line=bottom,
            width=\columnwidth,
            xmin=0, xmax=24,
            tick label style={font=\footnotesize},
            ymajorgrids=true,
            y=0.3mm,
            boxplot/draw direction=y,
            xtick={1,2,3,4,5,6,7,8,9,10,11,12,13,14,15,16,17,18,19,20,21,22,23},
            xticklabels={},
            ymin=0,ymax=60,
            ytick={0,10,20,30,40,50,60},
            ylabel={Height error [mm]},
            label style={font=\footnotesize},
        ]
        \addplot+[boxplot, boxplot/draw position=1,mark=*, mark options={white,scale=0.5},boxplot/box extend=0.5] table[y=id1]{\IODEHtransparent};
        \addplot+[boxplot, boxplot/draw position=2,mark=*, mark options={white,scale=0.5},boxplot/box extend=0.5] table[y=id7]{\IODEHtransparent};
        \addplot+[boxplot, boxplot/draw position=3,mark=*, mark options={white,scale=0.5},boxplot/box extend=0.5] table[y=id6]{\IODEHtransparent};
        \addplot+[boxplot, boxplot/draw position=4,mark=*, mark options={white,scale=0.5},boxplot/box extend=0.5] table[y=id17]{\IODEHtransparent};
        \addplot+[boxplot, boxplot/draw position=5,mark=*, mark options={white,scale=0.5},boxplot/box extend=0.5] table[y=id9]{\IODEHtransparent};
        \addplot+[boxplot, boxplot/draw position=7,mark=*, mark options={white,scale=0.5},boxplot/box extend=0.5] table[y=id13]{\IODEHtransparent};
        \addplot+[boxplot, boxplot/draw position=8,mark=*, mark options={white,scale=0.5},boxplot/box extend=0.5] table[y=id16]{\IODEHtransparent};
        \addplot+[boxplot, boxplot/draw position=9,mark=*, mark options={white,scale=0.5},boxplot/box extend=0.5] table[y=id8]{\IODEHtransparent};
         \addplot+[boxplot, boxplot/draw position=10,mark=*, mark options={white,scale=0.5},boxplot/box extend=0.5] table[y=id19]{\IODEHtransparent};
         \addplot+[boxplot, boxplot/draw position=11,mark=*, mark options={white,scale=0.5},boxplot/box extend=0.5] table[y=id20]{\IODEHtransparent};
        \addplot+[boxplot, boxplot/draw position=12,mark=*, mark options={white,scale=0.5},boxplot/box extend=0.5] table[y=id21]{\IODEHtransparent};
        \addplot+[boxplot, boxplot/draw position=13,mark=*, mark options={white,scale=0.5},boxplot/box extend=0.5] table[y=id22]{\IODEHtransparent};
        \addplot+[boxplot, boxplot/draw position=14,mark=*, mark options={white,scale=0.5},boxplot/box extend=0.5] table[y=id2]{\IODEHtranslucid};
        \addplot+[boxplot, boxplot/draw position=15,mark=*, mark options={white,scale=0.5},boxplot/box extend=0.5] table[y=id3]{\IODEHtranslucid};
        \addplot+[boxplot, boxplot/draw position=16,mark=*, mark options={white,scale=0.5},boxplot/box extend=0.5] table[y=id4]{\IODEHtranslucid};
        \addplot+[boxplot, boxplot/draw position=17,mark=*, mark options={white,scale=0.5},boxplot/box extend=0.5] table[y=id5]{\IODEHtranslucid};
        \addplot+[boxplot, boxplot/draw position=18,mark=*, mark options={white,scale=0.5},boxplot/box extend=0.5] table[y=id23]{\IODEHtranslucid};
        \addplot+[boxplot, boxplot/draw position=19,mark=*, mark options={white,scale=0.5},boxplot/box extend=0.5] table[y=id12]{\IODEHtranslucid};
         \addplot+[boxplot, boxplot/draw position=20,mark=*, mark options={white,scale=0.5},boxplot/box extend=0.5] table[y=id18]{\IODEHopaque};
        \addplot+[boxplot, boxplot/draw position=21,mark=*, mark options={white,scale=0.5},boxplot/box extend=0.5] table[y=id14]{\IODEHopaque};
        \addplot+[boxplot, boxplot/draw position=22,mark=*, mark options={white,scale=0.5},boxplot/box extend=0.5] table[y=id15]{\IODEHopaque};
        \addplot+[boxplot, boxplot/draw position=23,mark=*, mark options={white,scale=0.5},boxplot/box extend=0.5] table[y=id11]{\IODEHopaque};
        \end{axis}
        \end{tikzpicture}
        \begin{tikzpicture}
        \begin{axis}[
            axis x line=bottom,
            width=\columnwidth,
            xmin=0, xmax=24,
            tick label style={font=\footnotesize},
            ymajorgrids=true,
            y=1mm,
            boxplot/draw direction=y,
            xtick={1,2,3,4,5,6,7,8,9,10,11,12,13,14,15,16,17,18,19,20,21,22,23},
            xticklabels={,2,,,5,,,8,,,11,,,14,,,17,,,20,,,23},
            ymin=0,ymax=20,
            ytick={0,4,8,12,16,20},
            ylabel={Width error [mm]},
            label style={font=\footnotesize},
            xlabel={Object ID},
        ]
        \addplot+[boxplot, boxplot/draw position=1,mark=*, mark options={white,scale=0.5},boxplot/box extend=0.5] table[y=id1]{\IODEWtransparent};
        \addplot+[boxplot, boxplot/draw position=2,mark=*, mark options={white,scale=0.5},boxplot/box extend=0.5] table[y=id7]{\IODEWtransparent};
        \addplot+[boxplot, boxplot/draw position=3,mark=*, mark options={white,scale=0.5},boxplot/box extend=0.5] table[y=id6]{\IODEWtransparent};
        \addplot+[boxplot, boxplot/draw position=4,mark=*, mark options={white,scale=0.5},boxplot/box extend=0.5] table[y=id17]{\IODEWtransparent};
        \addplot+[boxplot, boxplot/draw position=5,mark=*, mark options={white,scale=0.5},boxplot/box extend=0.5] table[y=id9]{\IODEWtransparent};
        \addplot+[boxplot, boxplot/draw position=7,mark=*, mark options={white,scale=0.5},boxplot/box extend=0.5] table[y=id13]{\IODEWtransparent};
        \addplot+[boxplot, boxplot/draw position=8,mark=*, mark options={white,scale=0.5},boxplot/box extend=0.5] table[y=id16]{\IODEWtransparent};
        \addplot+[boxplot, boxplot/draw position=9,mark=*, mark options={white,scale=0.5},boxplot/box extend=0.5] table[y=id8]{\IODEWtransparent};
        \addplot+[boxplot, boxplot/draw position=10,mark=*, mark options={white,scale=0.5},boxplot/box extend=0.5] table[y=id19]{\IODEWtransparent};
        \addplot+[boxplot, boxplot/draw position=11,mark=*, mark options={white,scale=0.5},boxplot/box extend=0.5] table[y=id20]{\IODEWtransparent};
        \addplot+[boxplot, boxplot/draw position=12,mark=*, mark options={white,scale=0.5},boxplot/box extend=0.5] table[y=id21]{\IODEWtransparent};
        \addplot+[boxplot, boxplot/draw position=13,mark=*, mark options={white,scale=0.5},boxplot/box extend=0.5] table[y=id22]{\IODEWtransparent};
        \addplot+[boxplot, boxplot/draw position=19,mark=*, mark options={white,scale=0.5},boxplot/box extend=0.5] table[y=id2]{\IODEWtranslucid};
        \addplot+[boxplot, boxplot/draw position=14,mark=*, mark options={white,scale=0.5},boxplot/box extend=0.5] table[y=id3]{\IODEWtranslucid};
        \addplot+[boxplot, boxplot/draw position=15,mark=*, mark options={white,scale=0.5},boxplot/box extend=0.5] table[y=id4]{\IODEWtranslucid};
        \addplot+[boxplot, boxplot/draw position=16,mark=*, mark options={white,scale=0.5},boxplot/box extend=0.5] table[y=id5]{\IODEWtranslucid};
        \addplot+[boxplot, boxplot/draw position=17,mark=*, mark options={white,scale=0.5},boxplot/box extend=0.5] table[y=id23]{\IODEWtranslucid};
        \addplot+[boxplot, boxplot/draw position=18,mark=*, mark options={white,scale=0.5},boxplot/box extend=0.5] table[y=id12]{\IODEWtranslucid};
        \addplot+[boxplot, boxplot/draw position=20,mark=*, mark options={white,scale=0.5},boxplot/box extend=0.5] table[y=id18]{\IODEWopaque};
        \addplot+[boxplot, boxplot/draw position=21,mark=*, mark options={white,scale=0.5},boxplot/box extend=0.5] table[y=id14]{\IODEWopaque};
        \addplot+[boxplot, boxplot/draw position=22,mark=*, mark options={white,scale=0.5},boxplot/box extend=0.5] table[y=id15]{\IODEWopaque};
        \addplot+[boxplot, boxplot/draw position=23,mark=*, mark options={white,scale=0.5},boxplot/box extend=0.5] table[y=id11]{\IODEWopaque};
        \end{axis}
        \end{tikzpicture}
        \vspace{-0.2cm}
    \caption{Localisation success ratio (LSR) of all methods and errors for each dimension using LoDE for each object of the CORSMAL Container dataset, across all backgrounds and lighting conditions. Note the different scale of the y-axis.
    Legend:
    \acrshort{nocs}~\cite{Wang2019CVPR_NOCS}~\protect\tikz \protect\draw[nocs,fill=nocs] (0,0) rectangle (1.ex,1.ex);,
    SegDD~\protect\tikz \protect\draw[segdd,fill=segdd] (0,0) rectangle (1.ex,1.ex);,
    \acrshort{lodeir}~\protect\tikz \protect\draw[iodenarrow,fill=iodenarrow] (0,0) rectangle (1.ex,1.ex); and
    LoDE~\protect\tikz \protect\draw[iode,fill=iode] (0,0) rectangle (1.ex,1.ex);.
    }
    \label{fig:IODE}
    \vspace{-10pt}
\end{figure}
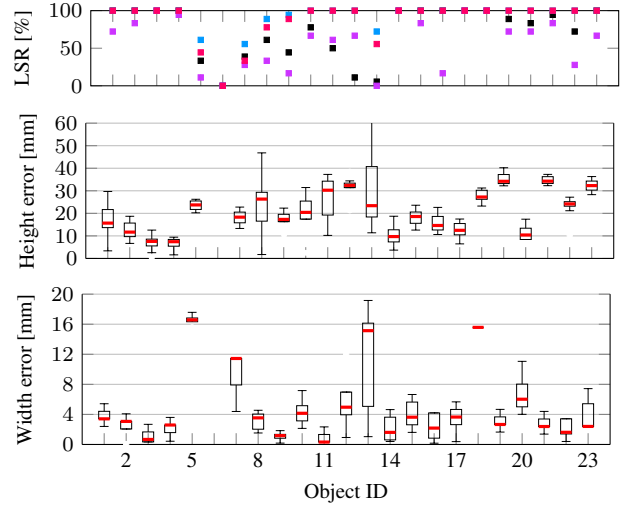

Fig.~\ref{fig:IODE} shows the statistics (median, min, max, 25 percentile and 75 percentile) of the dimensions error of our approach for each object across all the background and lighting variations.
LoDE accurately estimates the width of most of the objects with an error smaller than 20~mm and with small variations across the configurations. Objects 5 (juice glass), 7 (beer cup), 13 (champagne flute) and 18 (small white cup) are the least accurate cases, where the median error is larger than 10~mm. LoDE is less accurate in estimating the object height with the errors varying between $\sim$10~mm and $\sim$40~mm. This larger inaccuracy is due to the perspective on the image plane, as circumferences at lower/higher height than the real one are re-sampled with smaller radius to fit within the object masks. Objects 1 (bottle of water), 8 (plastic wine glass), 11 (rum glass) and 13 (champagne flute) show larger variations across configurations than other objects. As width and height are estimated independently, there is no correlation between the two dimensions.
While LoDE localises most of the objects across all the configurations (100\% LSR), there are some challenging cases, such as objects 5 (juice glass), 7 (beer cup) and 13 (champagne flute), where the LSR is below 60\%. Note that champagne flute is not localised by \acrshort{nocs} and \acrshort{lodeir}. 
The most challenging case for all methods is object 6 (diamond-shaped glass) that is never detected through the semantic segmentation due to the high level of transparency and the unusual shape. 
Moreover, \acrshort{nocs} and \acrshort{lodeir} obtain a lower LSR than LoDE for most of the transparent glasses/cups (\eg~objs.~5--13) and the small cups (objs.~18 and 22).  

\begin{figure}[t!]
    \centering
    \begin{tikzpicture}
        \begin{axis}[
            axis x line=bottom,
            width=0.55\columnwidth,
            xmin=0, xmax=15,
            ymajorgrids=true,
            y=0.22mm,
            boxplot/draw direction=y,
            xtick={1,2,3,4,6,7,8,9,11,12,13,14},
            xticklabels={NOCS, SegDD, \acrshort{lodeir}, \textbf{\acrshort{iode}}, NOCS, SegDD, \acrshort{lodeir}, \textbf{\acrshort{iode}}, NOCS, SegDD, \acrshort{lodeir}, \textbf{\acrshort{iode}}},
            xticklabel style = {rotate=75,anchor=east},
            ymin=0,ymax=120,
            ytick={0,20,40,60,80,100,120},
            ylabel={Height error [mm]},
            tick label style={font=\scriptsize},
            label style={font=\scriptsize},
            ticklabel style={font=\scriptsize},
            ylabel near ticks,
        ]
        \addplot+[boxplot, boxplot/draw position=1,mark=*, mark options={white,scale=0.5},boxplot/box extend=0.5] table[y=NOCS0]{\NOCSHtrans};
        \addplot+[boxplot, boxplot/draw position=2,mark=*, mark options={white,scale=0.5},boxplot/box extend=0.5] table[y=SegDD0]{\SegDDHtrans};
        \addplot+[boxplot, boxplot/draw position=3,mark=*, mark options={white,scale=0.5},boxplot/box extend=0.5] table[y=IODEnarrow0]{\IODEnarrowHtrans};
        \addplot+[boxplot, boxplot/draw position=4,mark=*, mark options={white,scale=0.5},boxplot/box extend=0.5] table[y=IODE0]{\IODEHtrans};
        \addplot+[boxplot, boxplot/draw position=6,mark=*, mark options={white,scale=0.5},boxplot/box extend=0.5] table[y=NOCS1]{\NOCSHtrans};
        \addplot+[boxplot, boxplot/draw position=7,mark=*, mark options={white,scale=0.5},boxplot/box extend=0.5] table[y=SegDD1]{\SegDDHtrans};
        \addplot+[boxplot, boxplot/draw position=8,mark=*, mark options={white,scale=0.5},boxplot/box extend=0.5] table[y=IODEnarrow1]{\IODEnarrowHtrans};
        \addplot+[boxplot, boxplot/draw position=9,mark=*, mark options={white,scale=0.5},boxplot/box extend=0.5] table[y=IODE1]{\IODEHtrans};
        \addplot+[boxplot, boxplot/draw position=11,mark=*, mark options={white,scale=0.5},boxplot/box extend=0.5] table[y=NOCS2]{\NOCSHtrans};
         \addplot+[boxplot, boxplot/draw position=12,mark=*, mark options={white,scale=0.5},boxplot/box extend=0.5] table[y=SegDD2]{\SegDDHtrans};
         \addplot+[boxplot, boxplot/draw position=13,mark=*, mark options={white,scale=0.5},boxplot/box extend=0.5] table[y=IODEnarrow2]{\IODEnarrowHtrans};
        \addplot+[boxplot, boxplot/draw position=14,mark=*, mark options={white,scale=0.5},boxplot/box extend=0.5] table[y=IODE2]{\IODEHtrans};
        \end{axis}
        \begin{axis}[
            axis x line=top,
            width=0.55\columnwidth,
            xmin=0, xmax=15,
            tick label style={font=\scriptsize},
            xtick={2.5,7.5,12.5},
            xticklabels={\textit{opaque},\textit{transl.},\textit{transp.}},
            typeset ticklabels with strut,
            label style={font=\scriptsize},
            ymin=0,ymax=120,
            yticklabels={},
            y=0.22mm,
        ]
        \end{axis}
        \end{tikzpicture}
        \begin{tikzpicture}
        \begin{axis}[
            axis x line=bottom,
            width=0.55\columnwidth,
            xmin=0, xmax=15,
            ymajorgrids=true,
            y=0.22mm,
            boxplot/draw direction=y,
            xtick={1,2,3,4,6,7,8,9,11,12,13,14},
            xticklabels={NOCS, SegDD, \acrshort{lodeir}, \textbf{\acrshort{iode}}, NOCS, SegDD, \acrshort{lodeir}, \textbf{\acrshort{iode}}, NOCS, SegDD, \acrshort{lodeir}, \textbf{\acrshort{iode}}},
            xticklabel style = {rotate=75,anchor=east},
            ymin=0,ymax=120,
            ytick={0,20,40,60,80,100,120},
            yticklabels={},
            ylabel={Width error [mm]},
            yticklabel pos=right,
            tick label style={font=\scriptsize},
            label style={font=\scriptsize},
            ticklabel style={font=\scriptsize},
            ylabel near ticks,
        ]
        \addplot+[boxplot, boxplot/draw position=1,mark=*, mark options={white,scale=0.5},boxplot/box extend=0.5] table[y=NOCS0]{\NOCSWtrans};
         \addplot+[boxplot, boxplot/draw position=2,mark=*, mark options={white,scale=0.5},boxplot/box extend=0.5] table[y=SegDD0]{\SegDDWtrans};
         \addplot+[boxplot, boxplot/draw position=3,mark=*, mark options={white,scale=0.5},boxplot/box extend=0.5] table[y=IODEnarrow0]{\IODEnarrowWtrans};
        \addplot+[boxplot, boxplot/draw position=4,mark=*, mark options={white,scale=0.5},boxplot/box extend=0.5] table[y=IODE0]{\IODEWtrans};
        \addplot+[boxplot, boxplot/draw position=6,mark=*, mark options={white,scale=0.5},boxplot/box extend=0.5] table[y=NOCS1]{\NOCSWtrans};
        \addplot+[boxplot, boxplot/draw position=7,mark=*, mark options={white,scale=0.5},boxplot/box extend=0.5] table[y=SegDD1]{\SegDDWtrans};
        \addplot+[boxplot, boxplot/draw position=8,mark=*, mark options={white,scale=0.5},boxplot/box extend=0.5] table[y=IODEnarrow1]{\IODEnarrowWtrans};
        \addplot+[boxplot, boxplot/draw position=9,mark=*, mark options={white,scale=0.5},boxplot/box extend=0.5] table[y=IODE1]{\IODEWtrans};
        \addplot+[boxplot, boxplot/draw position=11,mark=*, mark options={white,scale=0.5},boxplot/box extend=0.5] table[y=NOCS2]{\NOCSWtrans};
        \addplot+[boxplot, boxplot/draw position=12,mark=*, mark options={white,scale=0.5},boxplot/box extend=0.5] table[y=SegDD2]{\SegDDWtrans};
        \addplot+[boxplot, boxplot/draw position=13,mark=*, mark options={white,scale=0.5},boxplot/box extend=0.5] table[y=IODEnarrow2]{\IODEnarrowWtrans};
        \addplot+[boxplot, boxplot/draw position=14,mark=*, mark options={white,scale=0.5},boxplot/box extend=0.5] table[y=IODE2]{\IODEWtrans};
        \end{axis}
        \begin{axis}[
            axis x line=top,
            width=0.55\columnwidth,
            xmin=0, xmax=15,
            tick label style={font=\scriptsize},
            xtick={2.5,7.5,12.5},
            xticklabels={\textit{opaque},\textit{transl.},\textit{transp.}},
            label style={font=\scriptsize},
            ymin=0,ymax=120,
            ytick={0,20,40,60,80,100,120},
            yticklabels={},
            y=0.22mm,
            ylabel near ticks,
        ]
        \end{axis}
        \end{tikzpicture}
        \vspace{-0.3cm}
    \caption{Estimation error of height and width for opaque, translucent and transparent objects.}
    \label{fig:transparency}
\end{figure}
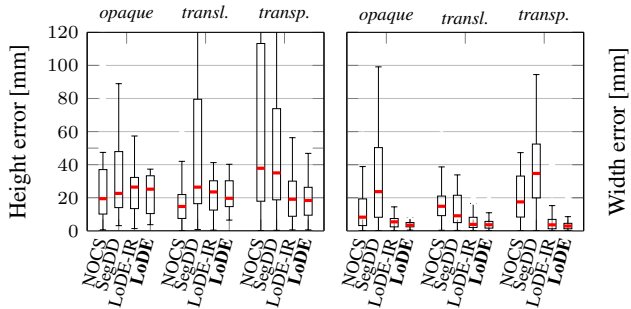

\pgfplotsset{
    every non boxed x axis/.style={},
    boxplot/every box/.style={solid,ultra thin,black},
    boxplot/every whisker/.style={solid,ultra thin,black},
    boxplot/every median/.style={solid,very thick, red},
}
\pgfplotstableread{acc.txt}\acc
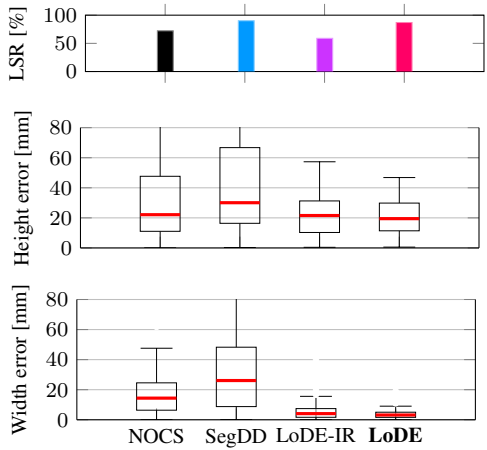
\begin{figure}[t!]
    \centering
    \begin{tikzpicture}
    \begin{axis}[
        ybar,
        width=.8\columnwidth,
        bar width=6pt,
        xmin=0,xmax=5,
        xtick={1,2,3,4},
        xticklabels={,,,},
        tick label style={font=\footnotesize},
        ylabel={LSR [\%]},
        ylabel near ticks,
        y=0.075mm,
        ymin=0, ymax=100,
        ytick={0,50,100},
        label style={font=\footnotesize},
    ]
    \addplot+[ybar, nocs, fill=nocs, draw opacity=0.5] coordinates {(1.4,72.22)};
    \addplot+[ybar, segdd, fill=segdd, draw opacity=0.5] coordinates {(2.15,90.10)};
    \addplot+[ybar, iodenarrow, fill=iodenarrow, draw opacity=0.5] coordinates {(2.87,58.94)};
    \addplot+[ybar, iode, fill=iode, draw opacity=0.5] coordinates {(3.6,86.96)};
    \end{axis}
    \end{tikzpicture}
    \begin{tikzpicture}
        \begin{axis}[
            axis x line=bottom,
            width=0.8\columnwidth,
            xmin=0, xmax=5,
            tick label style={font=\footnotesize},
            ymajorgrids=true,
            y=0.2mm,
            boxplot/draw direction=y,
            xtick={1,2,3,4},
            xticklabels={,,,},
            ymin=0,ymax=80,
            ytick={0,20,40,60,80},
            ylabel={Height error [mm]},
            label style={font=\footnotesize},
        ]
        \addplot+[boxplot, boxplot/draw position=1,mark=*, mark options={white,scale=0.5},boxplot/box extend=0.5] table[y=NOCSh]{\acc};
        \addplot+[boxplot, boxplot/draw position=2,mark=*, mark options={white,scale=0.5},boxplot/box extend=0.5] table[y=SegDDh]{\acc};
        \addplot+[boxplot, boxplot/draw position=3,mark=*, mark options={white,scale=0.5},boxplot/box extend=0.5] table[y=LoDEIRh]{\acc};
        \addplot+[boxplot, boxplot/draw position=4,mark=*, mark options={white,scale=0.5},boxplot/box extend=0.5] table[y=LoDEh]{\acc};
        \end{axis}
        \end{tikzpicture}
        \begin{tikzpicture}
        \begin{axis}[
            axis x line=bottom,
            width=0.8\columnwidth,
            xmin=0, xmax=5,
            tick label style={font=\footnotesize},
            ymajorgrids=true,
            y=0.2mm,
            boxplot/draw direction=y,
            xtick={1,2,3,4},
            xticklabels={NOCS,SegDD,LoDE-IR,\textbf{LoDE}},
            ymin=0,ymax=80,
            ytick={0,20,40,60,80},
            ylabel={Width error [mm]},
            label style={font=\footnotesize},
        ]
        \addplot+[boxplot, boxplot/draw position=1,mark=*, mark options={white,scale=0.5},boxplot/box extend=0.5] table[y=NOCSw]{\acc};
        \addplot+[boxplot, boxplot/draw position=2,mark=*, mark options={white,scale=0.5},boxplot/box extend=0.5] table[y=SegDDw]{\acc};
        \addplot+[boxplot, boxplot/draw position=3,mark=*, mark options={white,scale=0.5},boxplot/box extend=0.5] table[y=LoDEIRw]{\acc};
        \addplot+[boxplot, boxplot/draw position=4,mark=*, mark options={white,scale=0.5},boxplot/box extend=0.5] table[y=LoDEw]{\acc};
        \end{axis}
        \end{tikzpicture}
        \vspace{-0.2cm}
    \caption{Localisation success ratio (LSR) and dimension estimation error.
    Legend:
    \acrshort{nocs}~\cite{Wang2019CVPR_NOCS}~\protect\tikz \protect\draw[nocs,fill=nocs] (0,0) rectangle (1.ex,1.ex);,
    SegDD~\protect\tikz \protect\draw[segdd,fill=segdd] (0,0) rectangle (1.ex,1.ex);,
    \acrshort{lodeir}~\protect\tikz \protect\draw[iodenarrow,fill=iodenarrow] (0,0) rectangle (1.ex,1.ex); and
    \acrshort{iode}~\protect\tikz \protect\draw[iode,fill=iode] (0,0) rectangle (1.ex,1.ex);.
    }
    \label{fig:accvseff}
\end{figure}

Fig.~\ref{fig:transparency} compares the methods under varying degrees of transparency, such as opaque, translucent and transparent. The error is computed only for the cases where the object is successfully localised. As previously observed for LoDE, we can observe even here that all methods estimate the width more accurately than the height. The top-down perspective of the cameras makes the segmentation treat different parts of the object as one and consequently affects the height estimation when back-projecting in 3D via depth map or triangulation, or projecting for circumference verification. SegDD is more inaccurate for both translucent and transparent objects, with large variations especially in the height, due to the inaccuracies of the depth maps, while \acrshort{nocs} is inaccurate for transparent objects if localised. However, \acrshort{nocs} and SegDD are more accurate in estimating the height for opaque objects, while \acrshort{lodeir} and LoDE estimate the dimensions with a median error smaller than 30~mm despite the object transparency. 

Fig.~\ref{fig:accvseff} shows the success in localising the objects (LSR) and the error in estimating the height and width dimensions, across all the configurations. As previously observed, LoDE outperforms \acrshort{nocs} and SegDD obtaining 2.6~mm and 10.6~mm more accurate height estimations, and  11.2~mm and 22.9~mm more accurate width estimations comparing their medians, respectively, with a smaller standard deviation. LoDE also outperforms \acrshort{lodeir} in both height and width estimations. Furthermore, LoDE has a 25\% LSR higher than \acrshort{lodeir} at similar dimension error. 
Although both LoDE and SegDD uses Mask R-CNN, the LSR of LoDE is slightly lower than SegDD, as LoDE considers the two views simultaneously, while SegDD works on each view individually. 

\begin{figure}[t!]
    \centering
    \footnotesize
    \setlength\tabcolsep{0.3pt}
    \renewcommand{\arraystretch}{0.5}
    \begin{tabular}{cccccccc}
        & Obj.~22 & Obj.~20 & Obj.~15 & Obj.~12 & Obj.~12 & Obj.~7 & Obj.~5\\
        \raisebox{0.75cm}{\rotatebox[origin=c]{90}{NOCS}}  &
        \includegraphics[height=0.19\columnwidth, trim={580px 200px 480px 225px} ,clip]{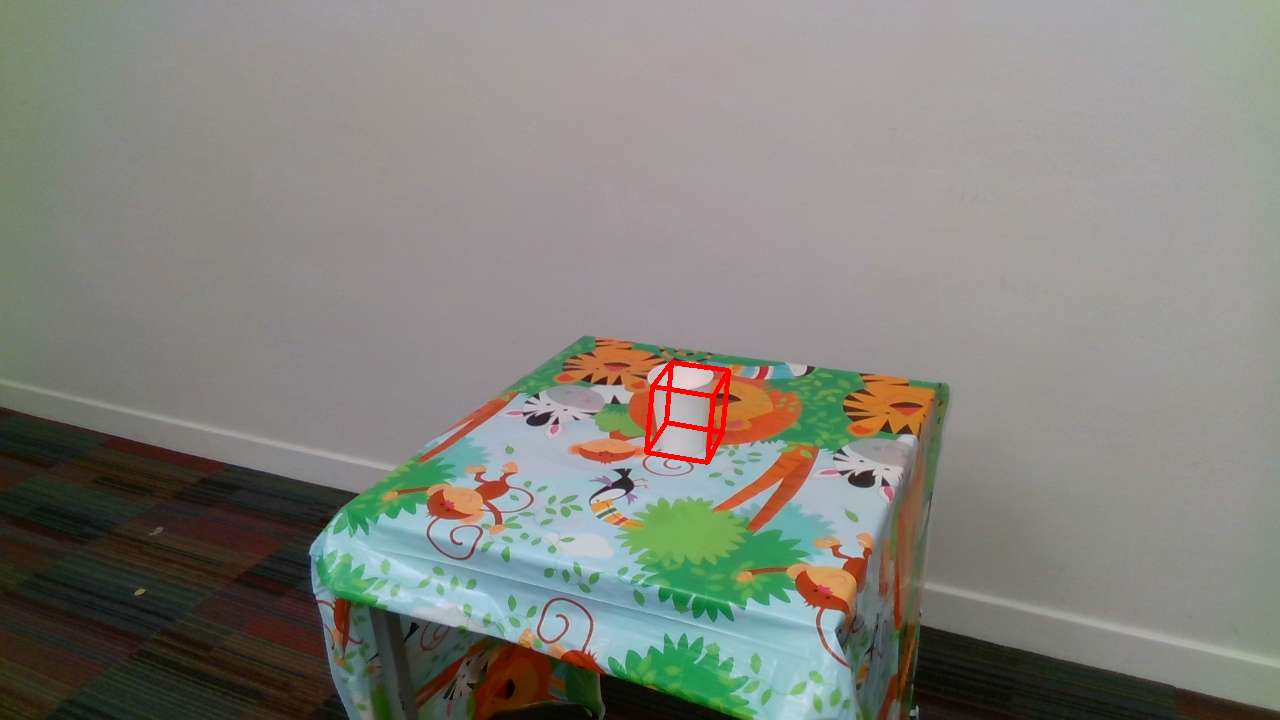} &
        \includegraphics[height=0.19\columnwidth, trim={650px 170px 420px 210px} ,clip]{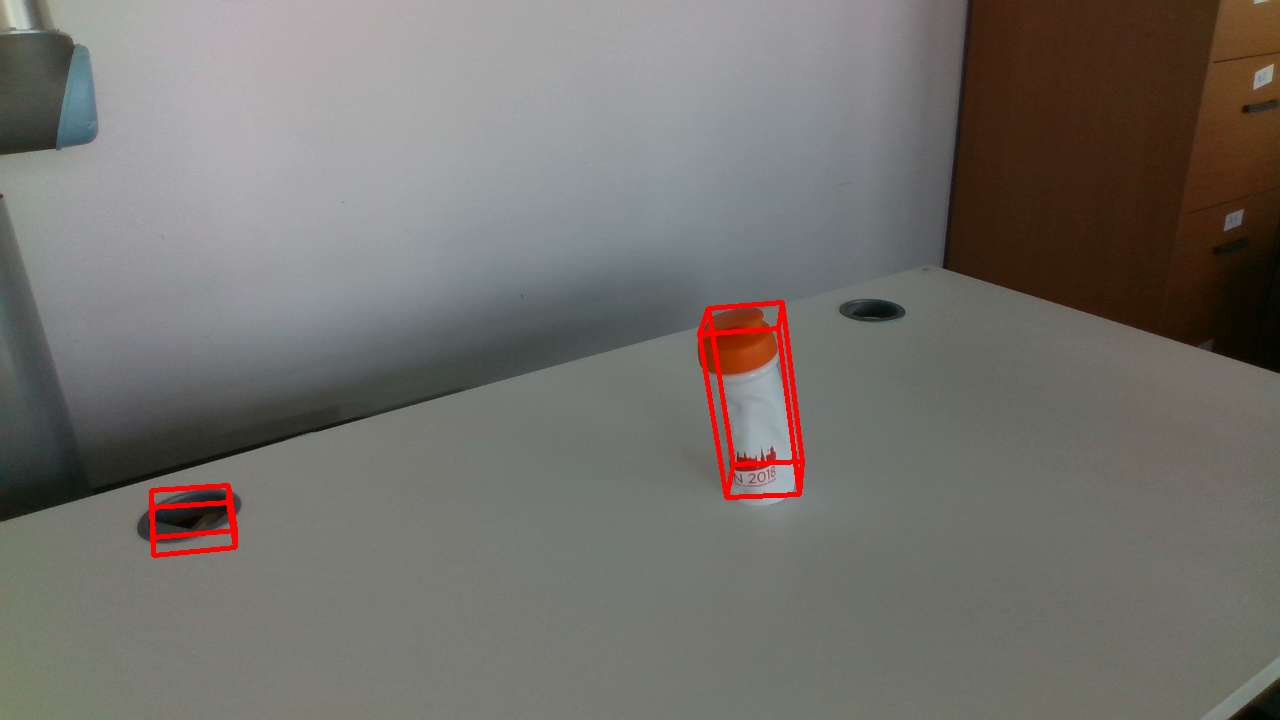} &
        \includegraphics[height=0.19\columnwidth, trim={550px 220px 475px  150px} ,clip]{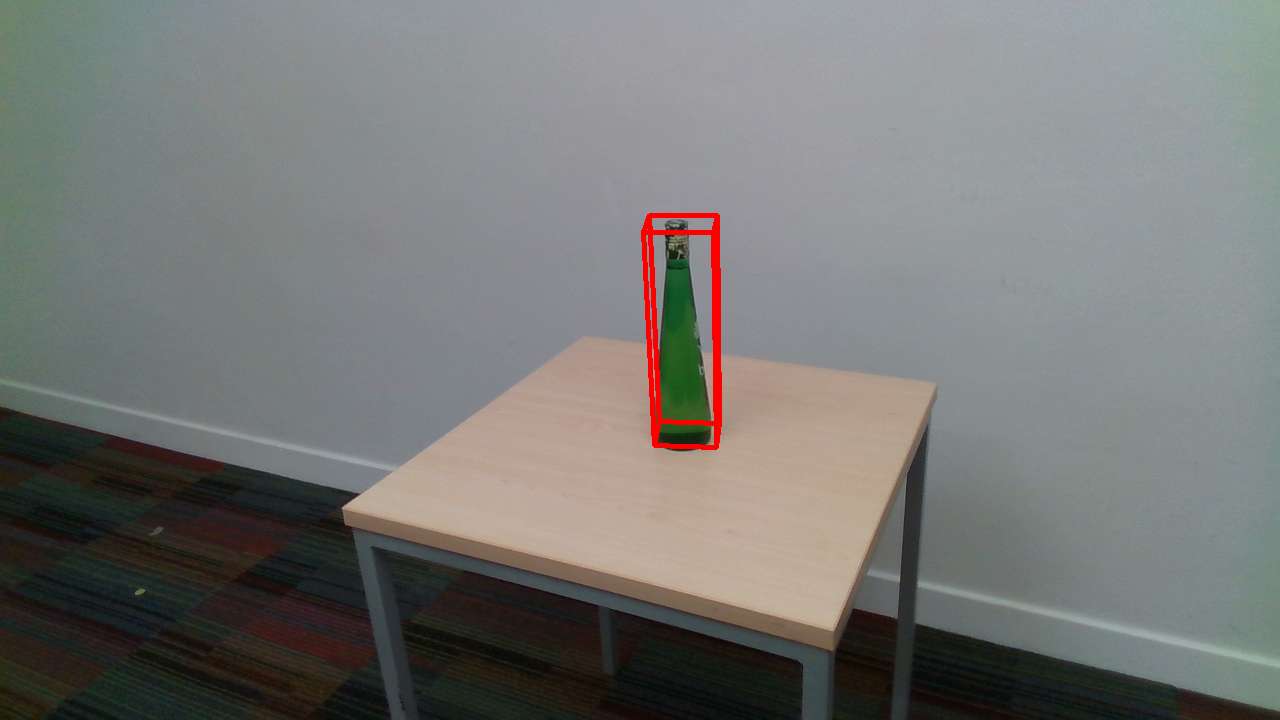} &
        \includegraphics[height=0.19\columnwidth, trim={605px 220px 530px 250px} ,clip]{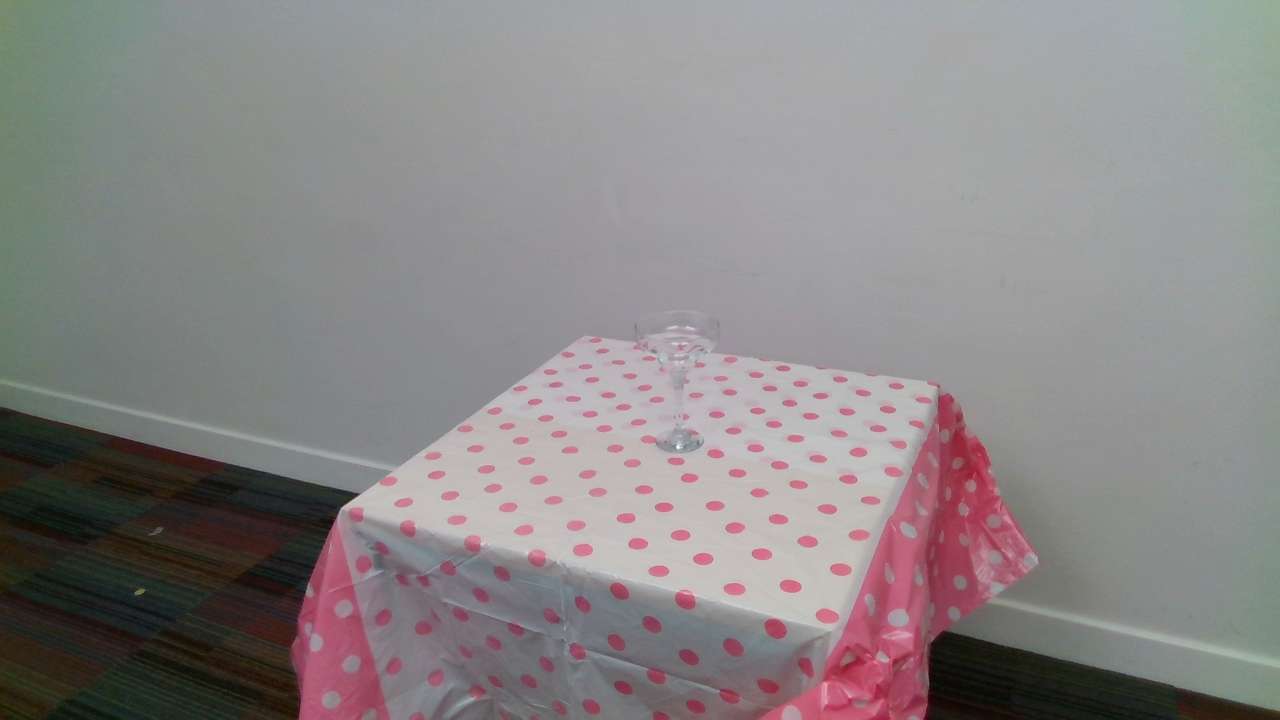} &
        \includegraphics[height=0.19\columnwidth, trim={650px 150px 375px 200px} ,clip]{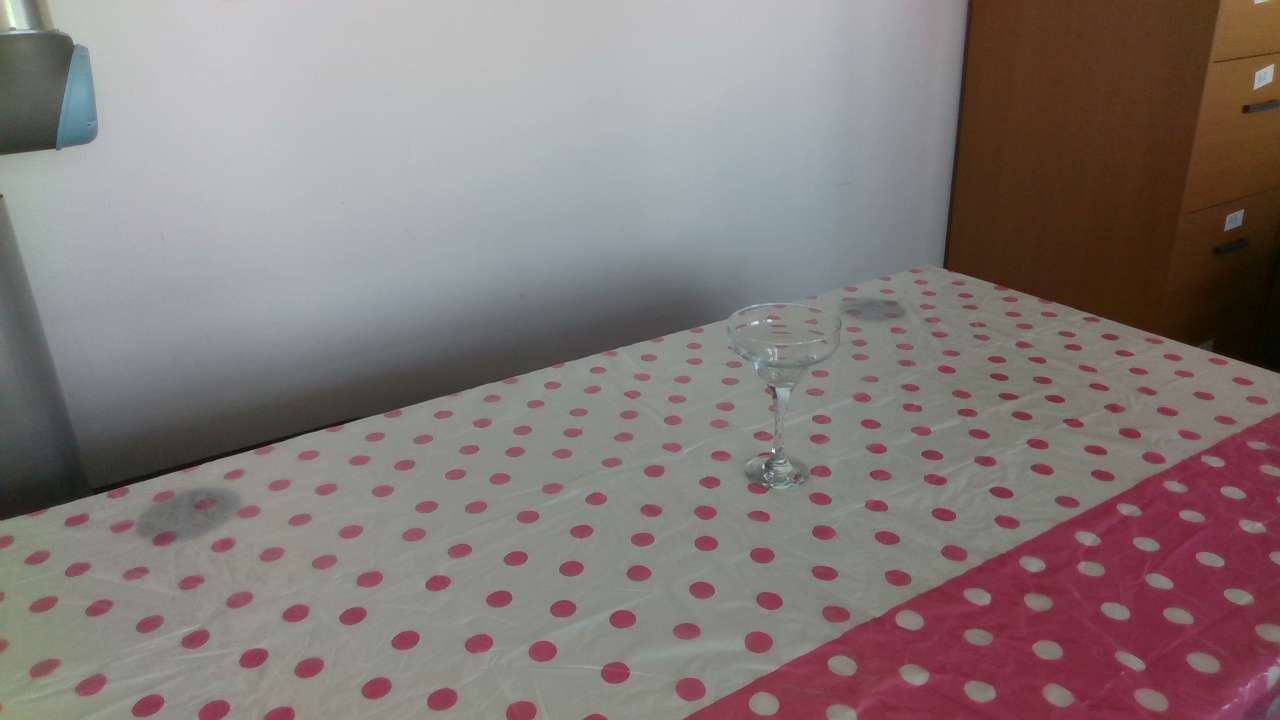} &
        \includegraphics[height=0.19\columnwidth, trim={630px 160px 405px 225px} ,clip]{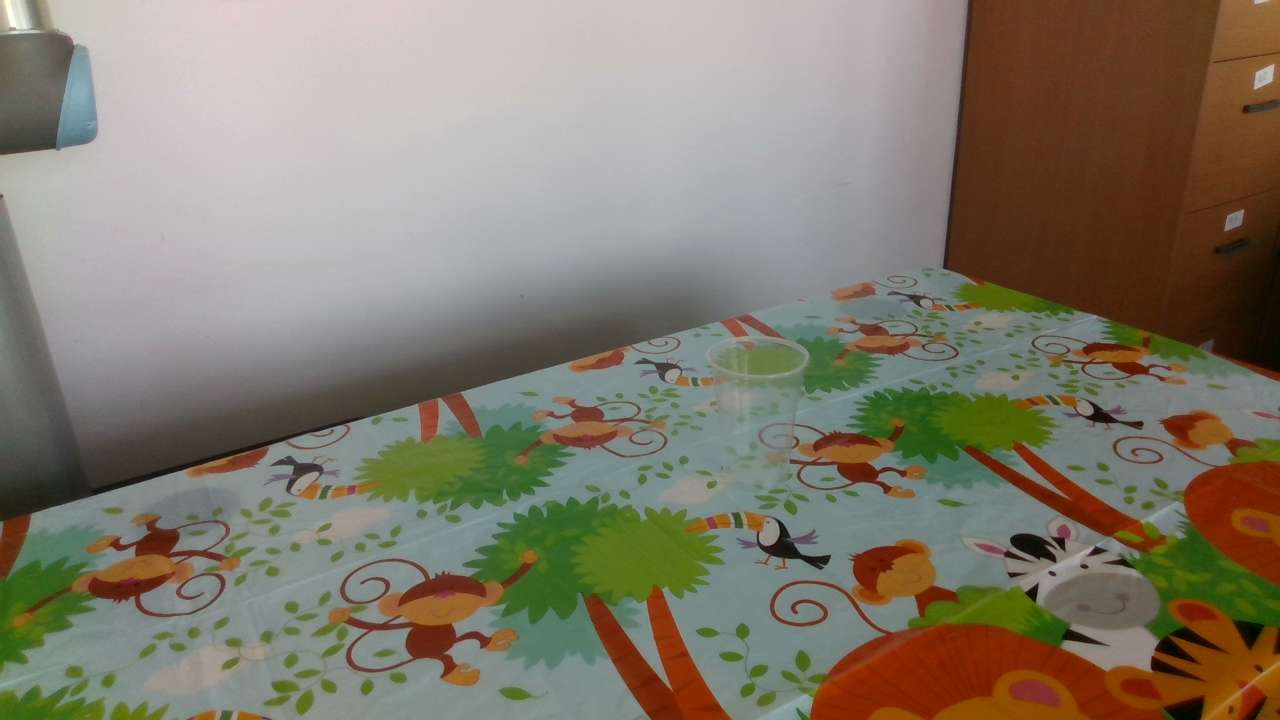} &
        \includegraphics[height=0.19\columnwidth, trim={650px 150px 400px 225px} ,clip]{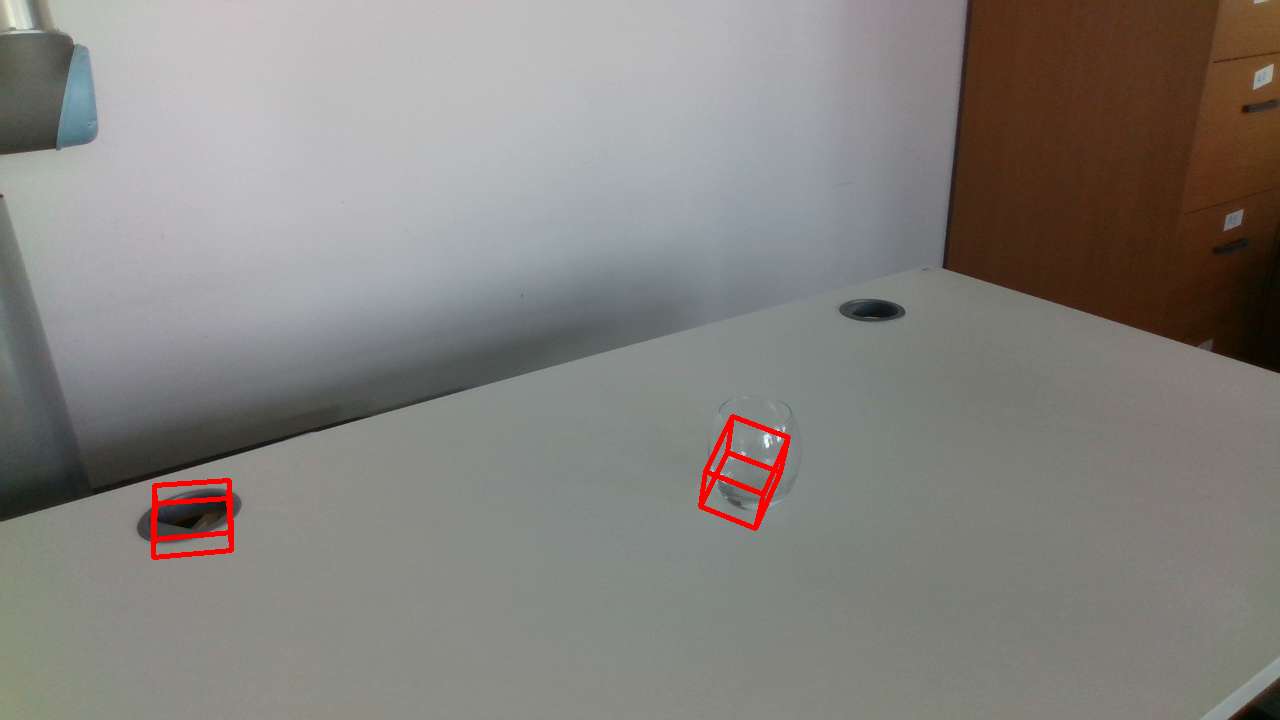} \\
        \raisebox{0.75cm}{\rotatebox[origin=c]{90}{SegDD}}  &
        \includegraphics[height=0.19\columnwidth, trim={580px 200px 480px 225px} ,clip]{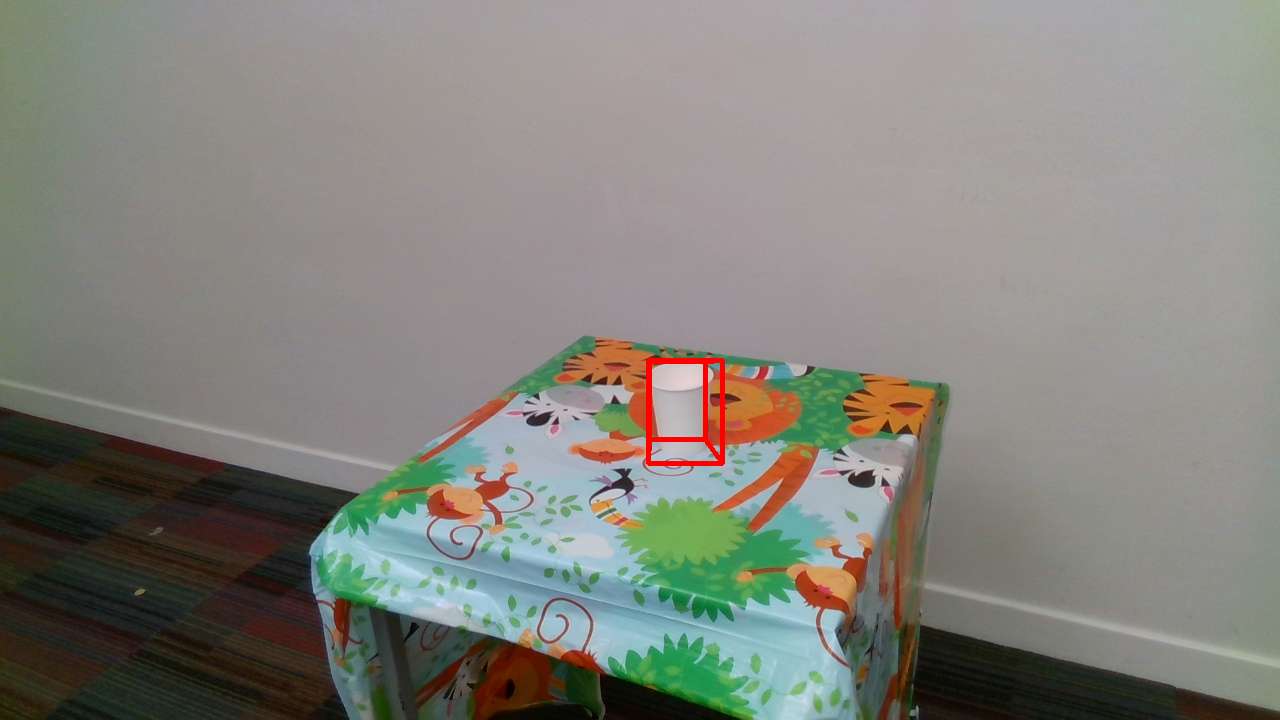} &
        \includegraphics[height=0.19\columnwidth, trim={650px 170px 420px 210px} ,clip]{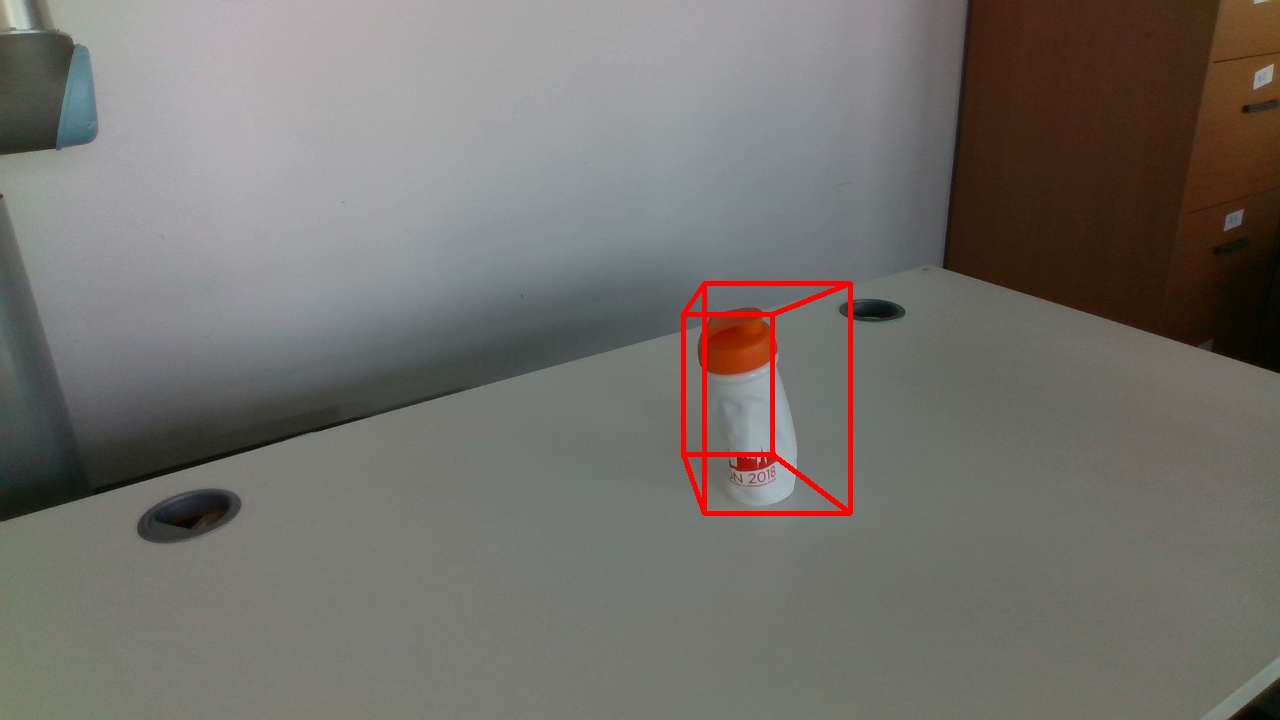} &
        \includegraphics[height=0.19\columnwidth, trim={550px 220px 475px  150px} ,clip]{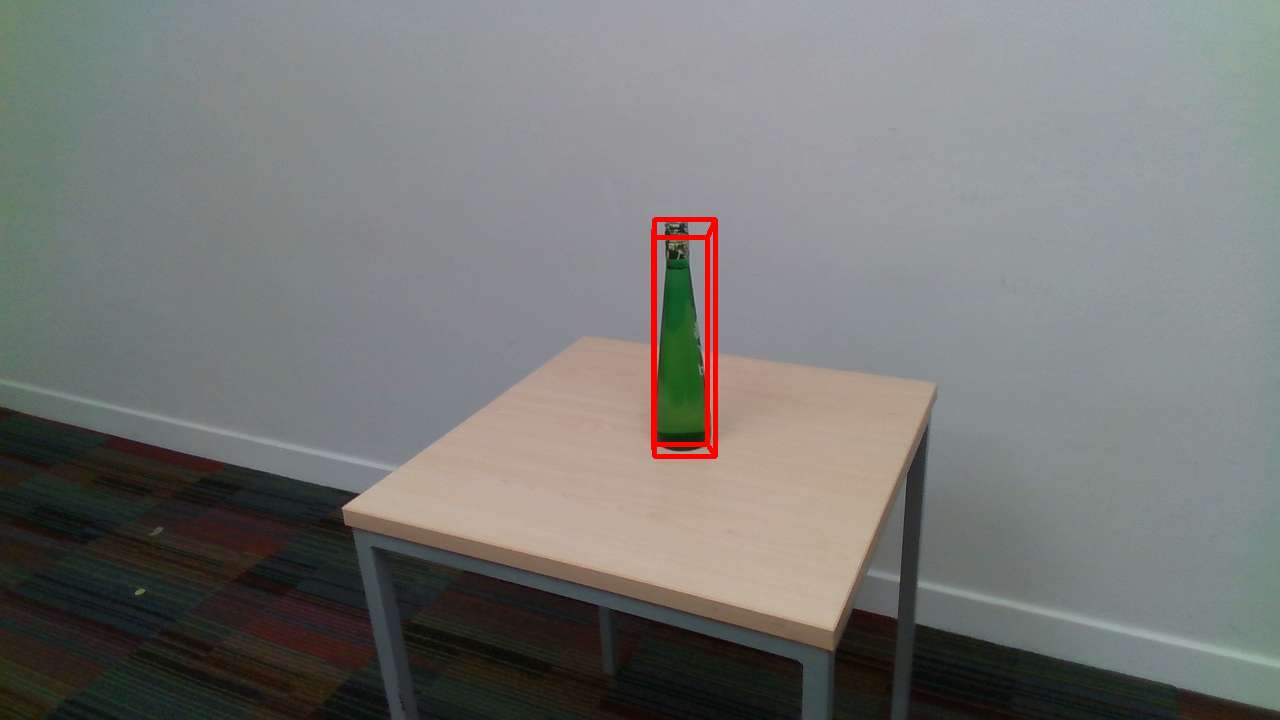} &
        \includegraphics[height=0.19\columnwidth, trim={605px 220px 530px 250px} ,clip]{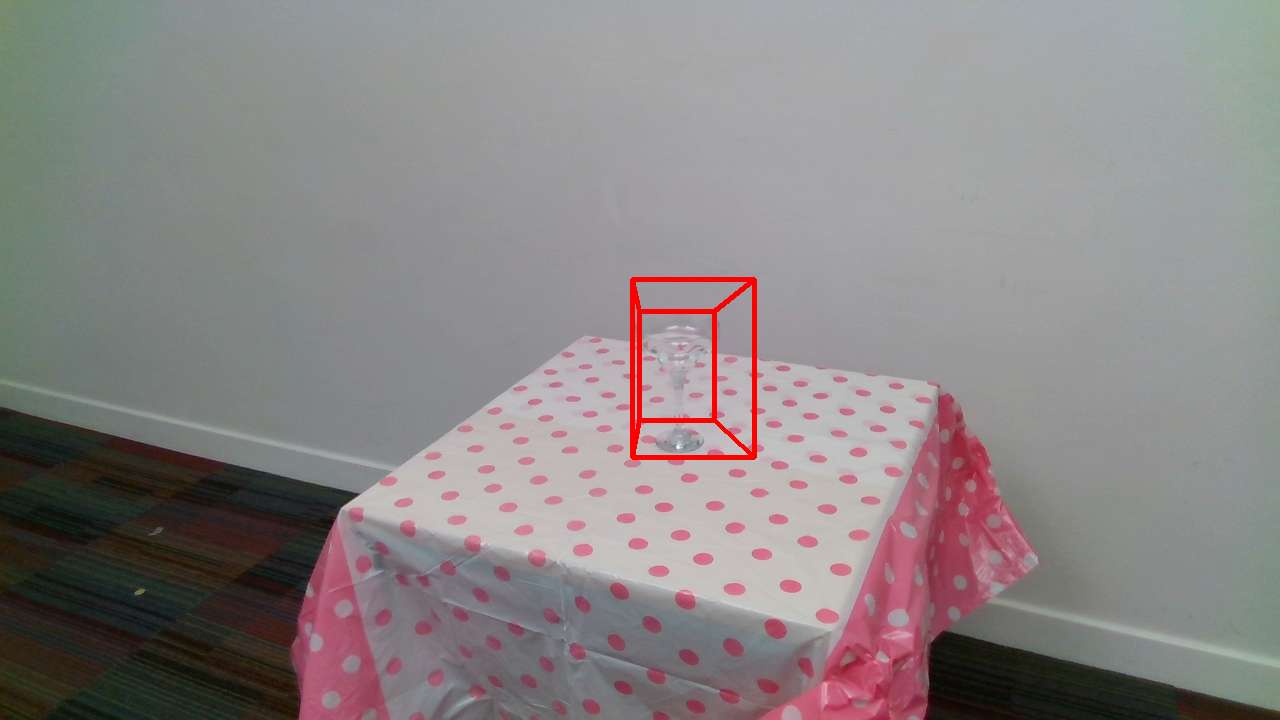} &
        \includegraphics[height=0.19\columnwidth, trim={650px 150px 375px 200px} ,clip]{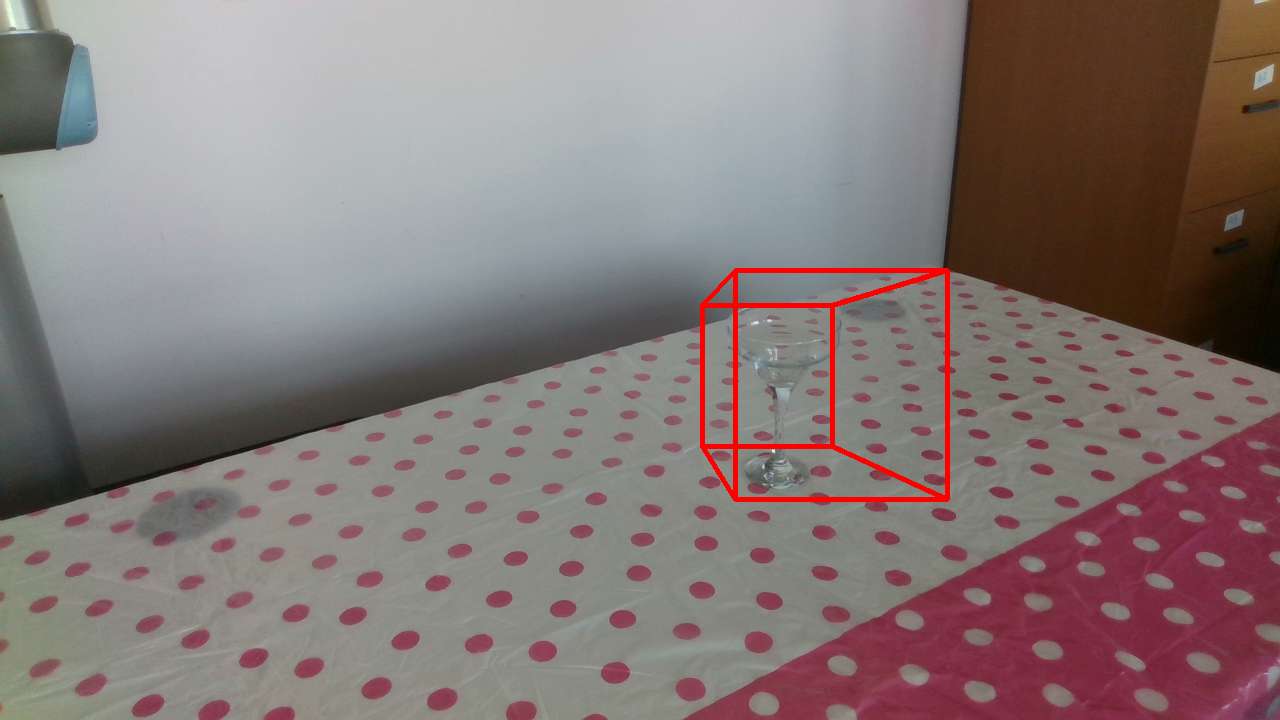} &
        \includegraphics[height=0.19\columnwidth, trim={630px 160px 405px 225px} ,clip]{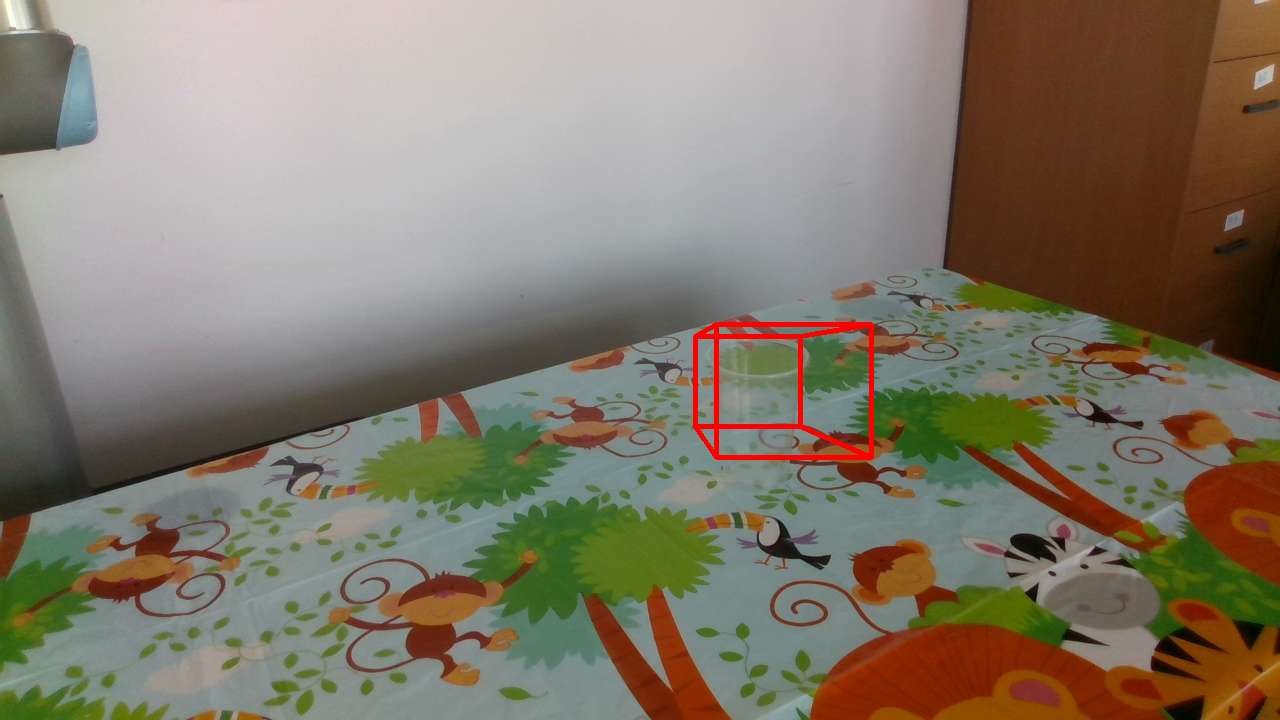} &
        \includegraphics[height=0.19\columnwidth, trim={650px 150px 400px 225px} ,clip]{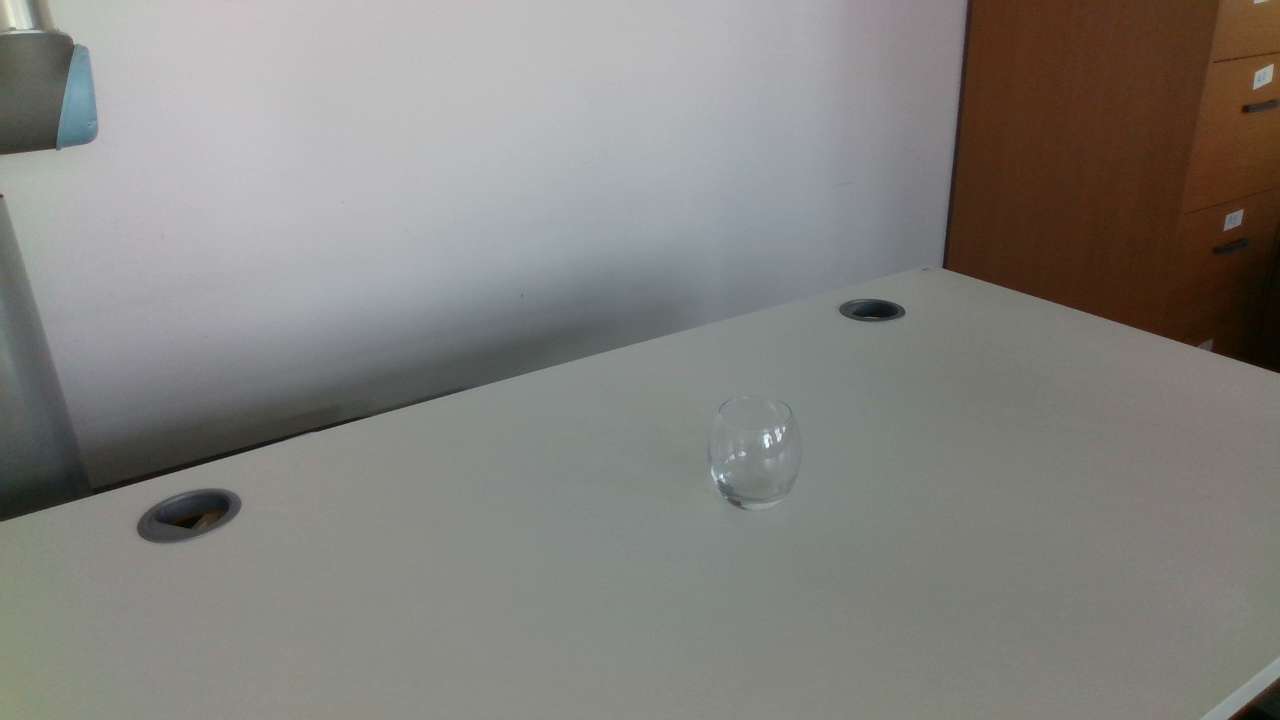} \\
        \raisebox{0.75cm}{\rotatebox[origin=c]{90}{LoDE-IR}}  &
        \includegraphics[height=0.19\columnwidth, trim={580px 275px 1845px 260px} ,clip]{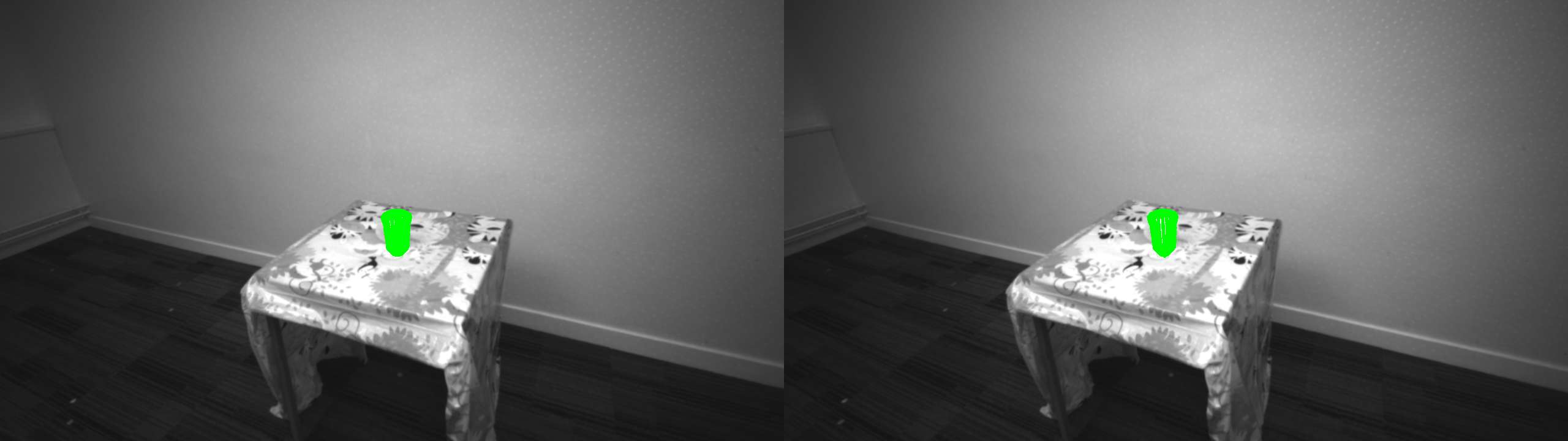} &
        \includegraphics[height=0.19\columnwidth, trim={620px 250px 520px 250px} ,clip]{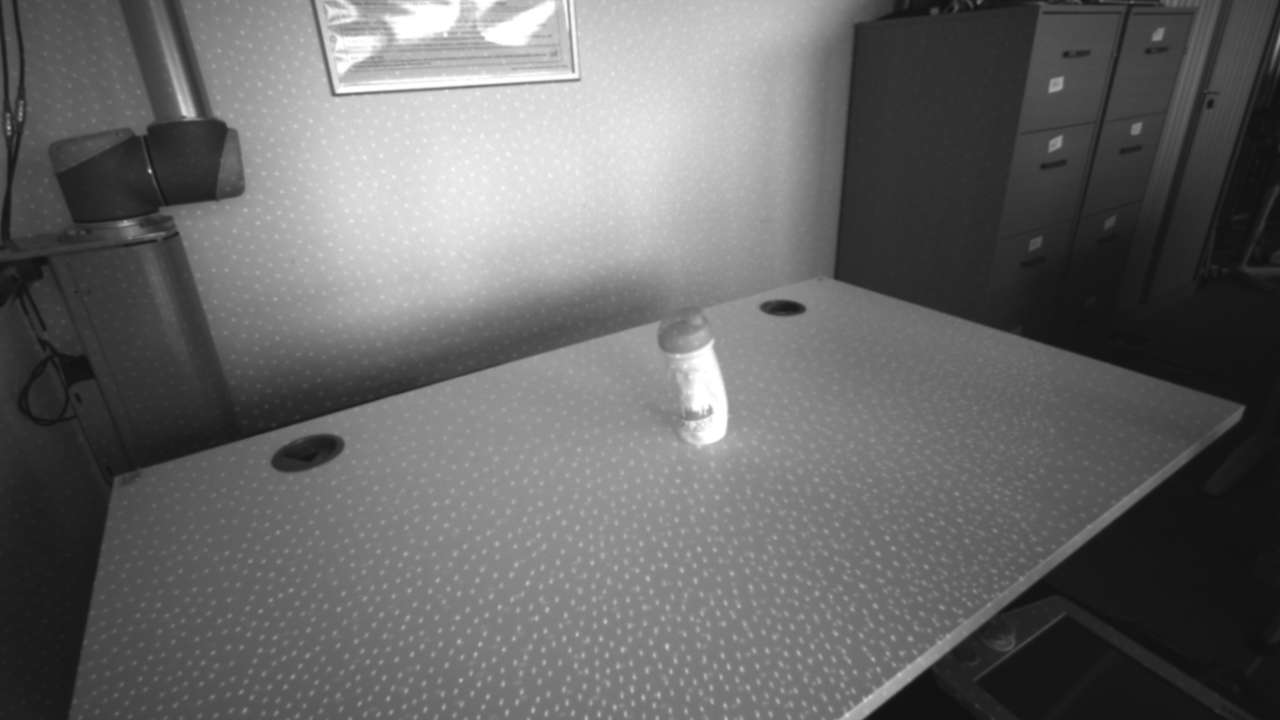} &
        \includegraphics[height=0.19\columnwidth, trim={575px 300px 560px 220px} ,clip]{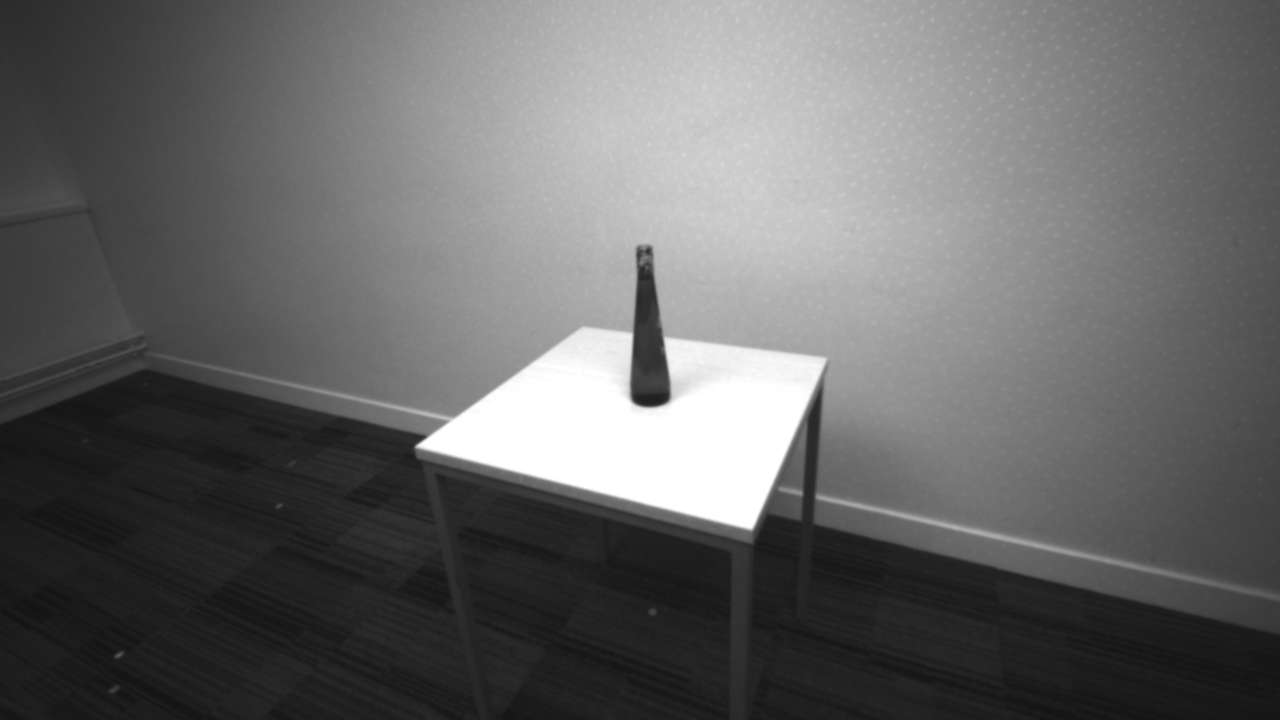} &
        \includegraphics[height=0.19\columnwidth, trim={600px 295px 1865px 260px} ,clip]{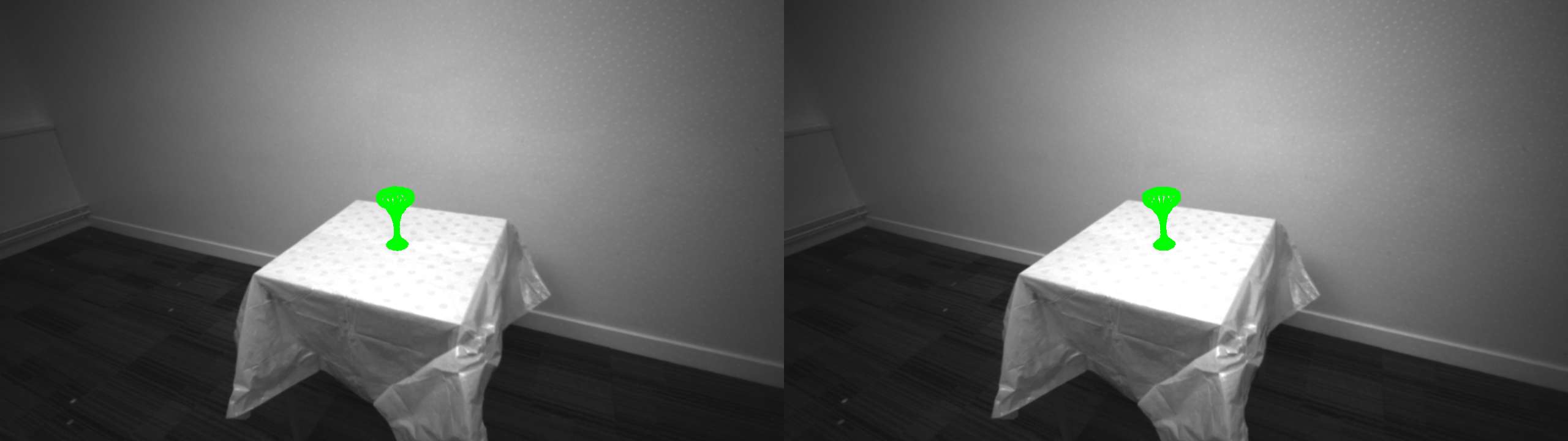} &
        \includegraphics[height=0.19\columnwidth, trim={643px 250px 1768px 250px} ,clip]{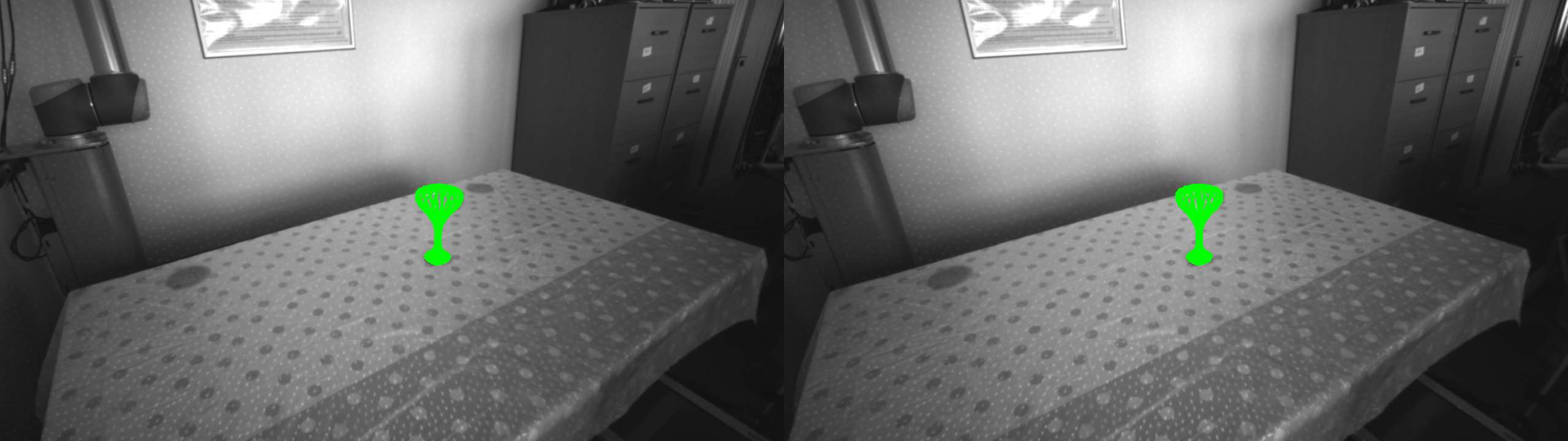} &
        \includegraphics[height=0.19\columnwidth, trim={630px 260px 1780px 250px} ,clip]{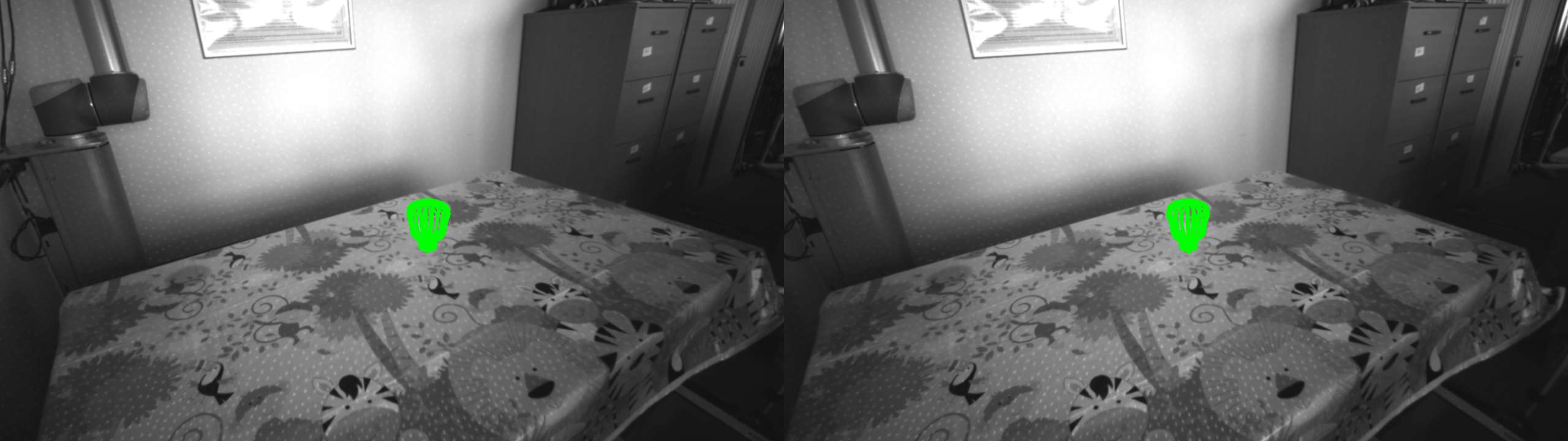} &
        \includegraphics[height=0.19\columnwidth, trim={625px 250px 500px 250px} ,clip]{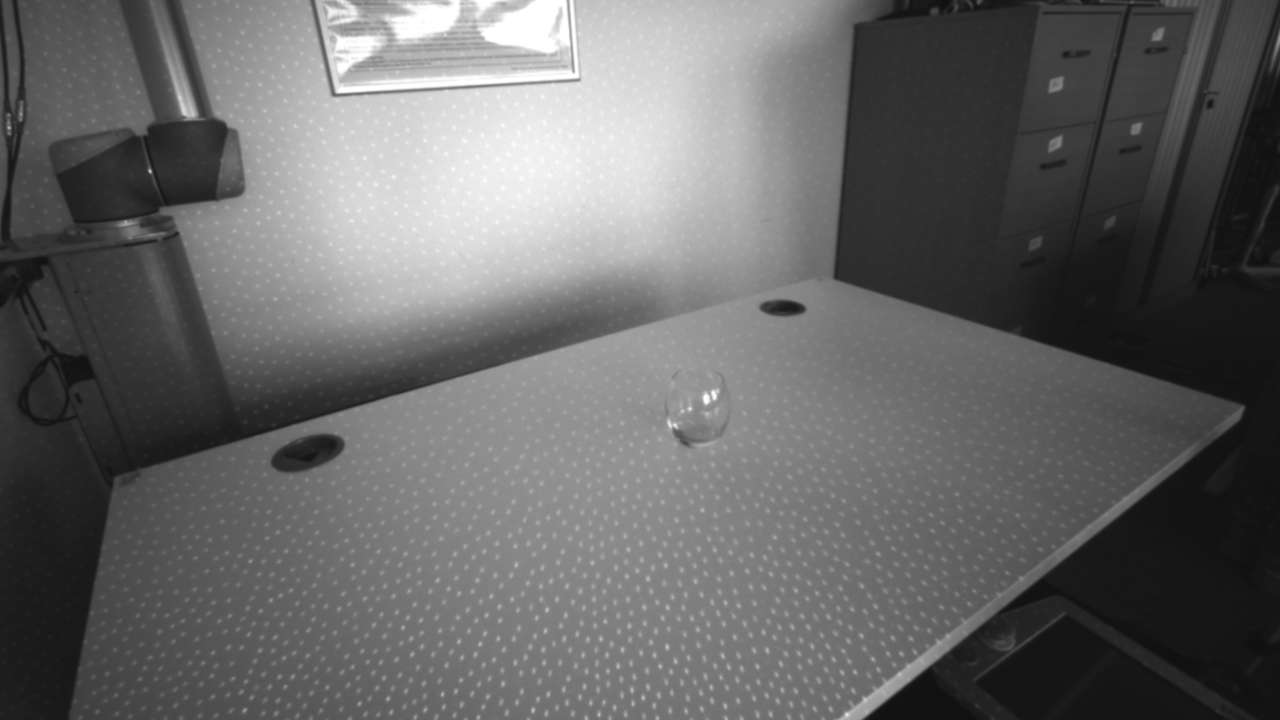} \\
        \raisebox{0.75cm}{\rotatebox[origin=c]{90}{\textbf{LoDE}}}  &
        \includegraphics[height=0.19\columnwidth, trim={580px 200px 1760px 225px} ,clip]{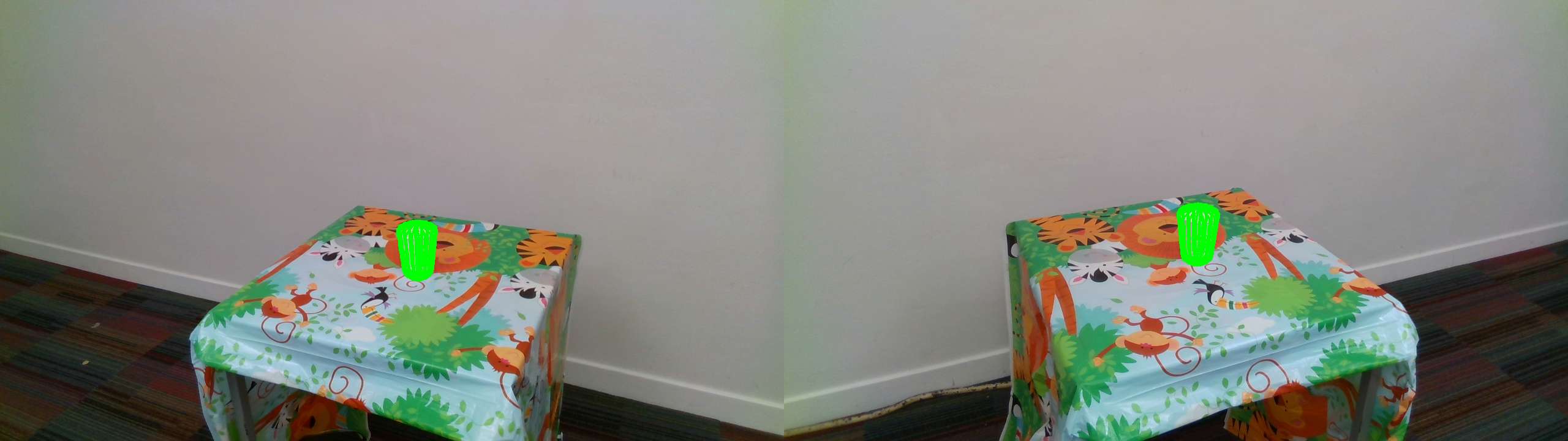} &
        \includegraphics[height=0.19\columnwidth, trim={650px 170px 1700px 210px} ,clip]{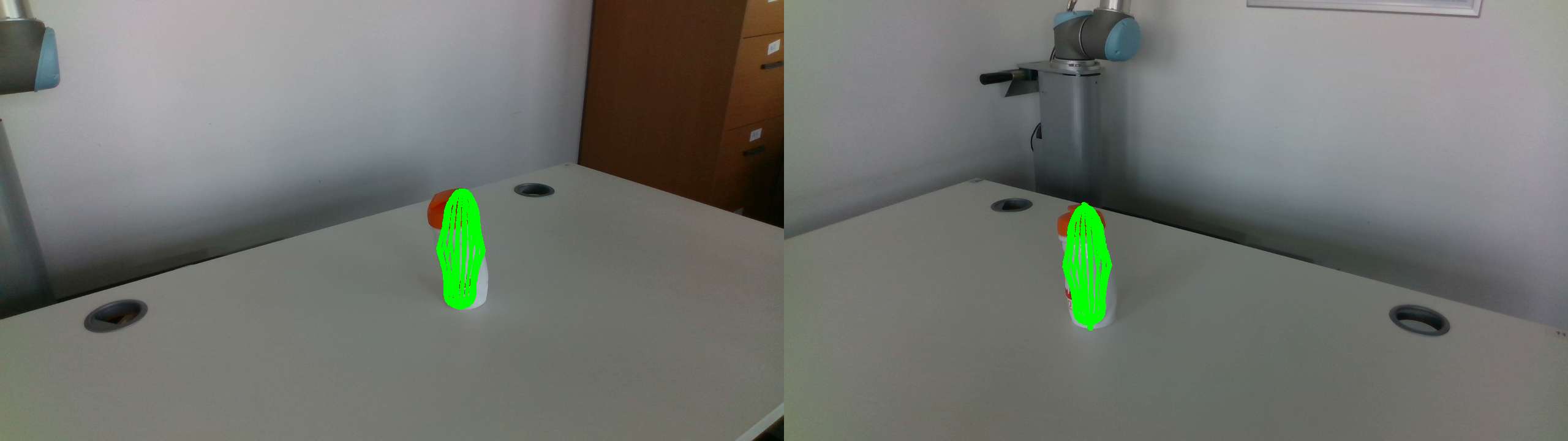} &
        \includegraphics[height=0.19\columnwidth, trim={550px 220px 1755px  150px} ,clip]{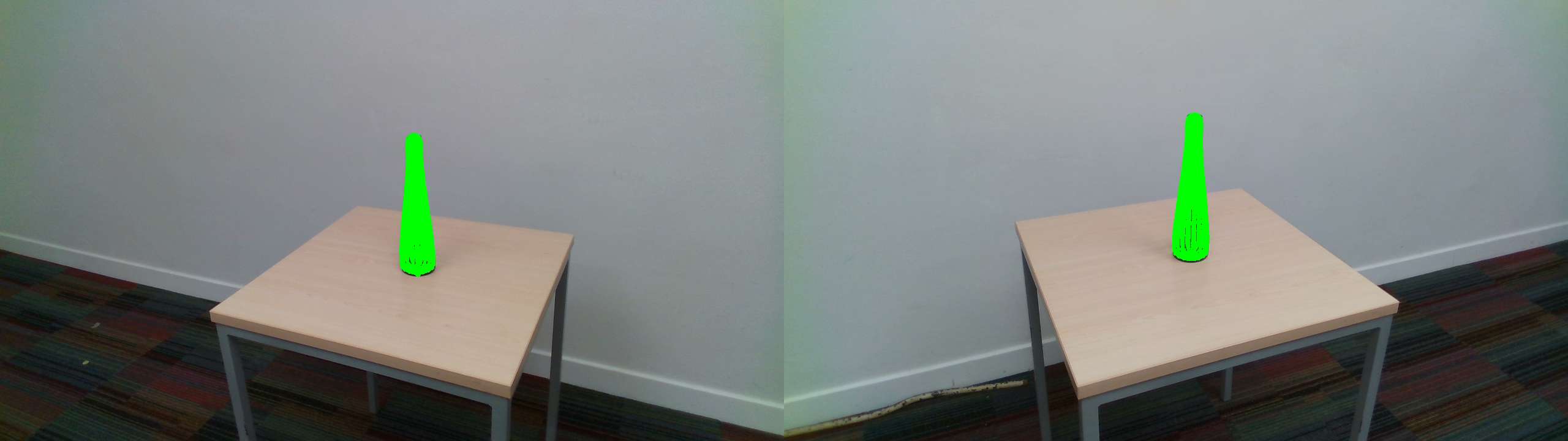} &
        \includegraphics[height=0.19\columnwidth, trim={605px 220px 1810px 250px} ,clip]{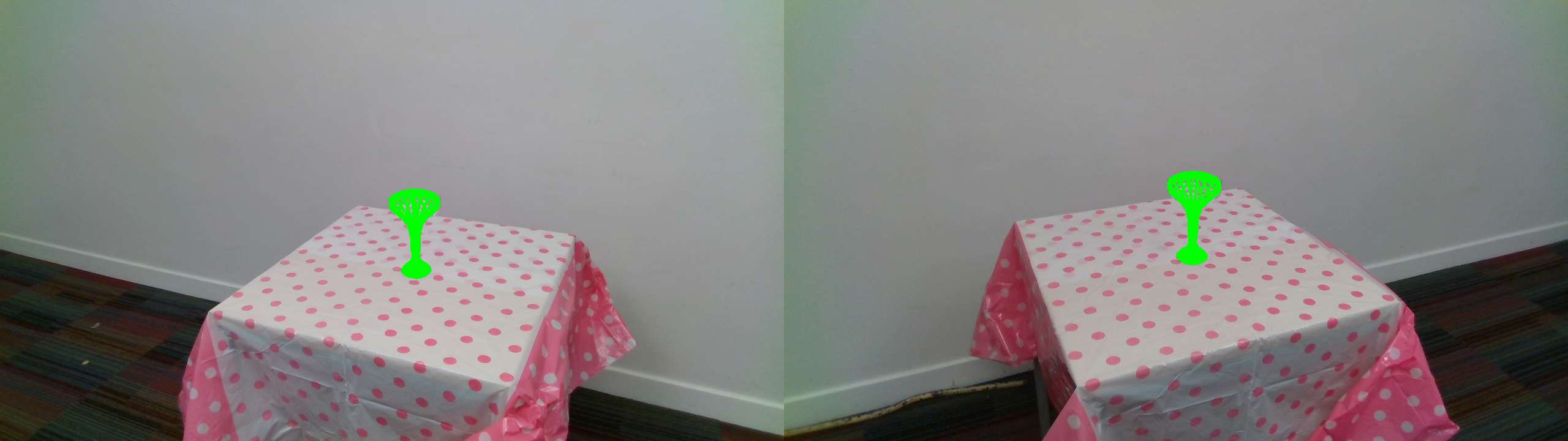} &
        \includegraphics[height=0.19\columnwidth, trim={650px 150px 1655px 200px} ,clip]{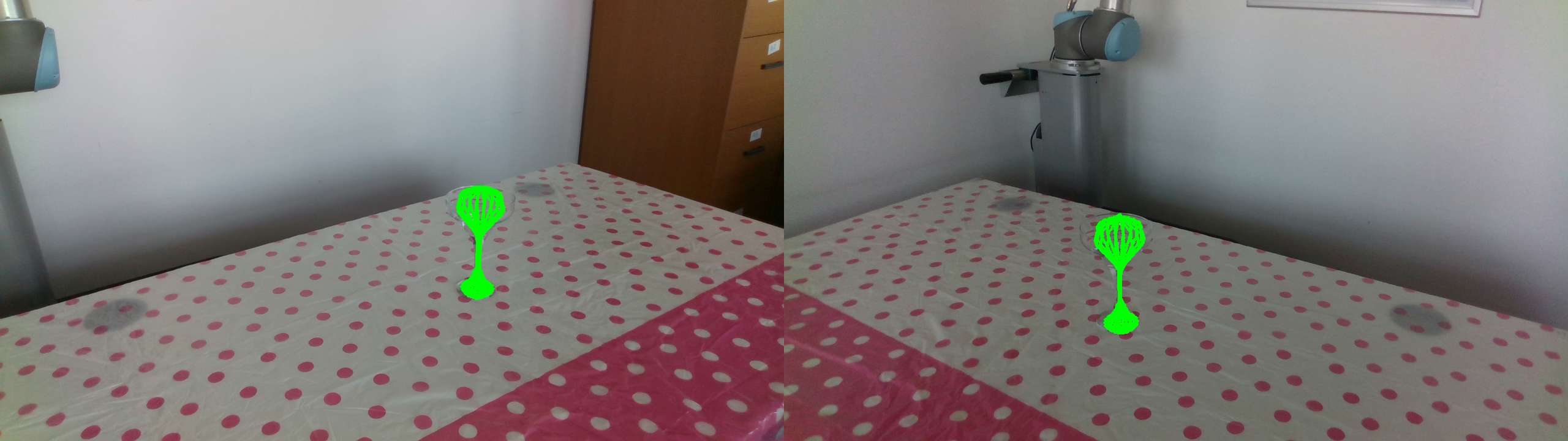} &
        \includegraphics[height=0.19\columnwidth, trim={630px 160px 1685px 225px} ,clip]{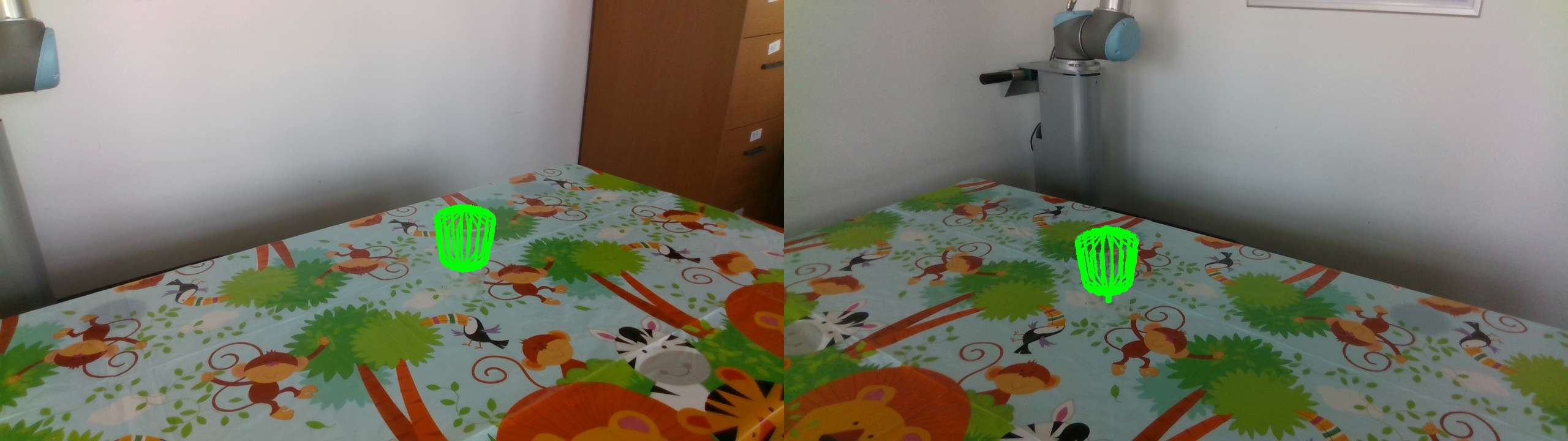}&
        \includegraphics[height=0.19\columnwidth, trim={650px 150px 400px 225px} ,clip]{147.jpg}  \\
         \\
    \end{tabular}
    \vspace{-0.2cm}
    \caption{Sample results for objects with varying transparency, backgrounds and lighting.
    Fourth and fifth columns correspond to the same object and background but different lighting (artificial and natural, respectively). KEY -- Obj.: object.
    }
    \label{fig:qualitexam}
    \vspace{-15pt}
\end{figure}

Fig.~\ref{fig:qualitexam} compares the results for one opaque and one transparent cup (objs.~22 and 7), one opaque and one translucent bottle (objs.~20 and 15), and two transparent drinking glasses (objs.~12 and 5) under different backgrounds and lighting conditions. All methods accurately estimate the dimensions of the opaque cup (obj.~22).
While SegDD, \acrshort{lodeir} and LoDE fail to localise obj.~5 (juice glass) under natural light, the bounding box estimated by \acrshort{nocs} is inaccurate. Moreover, \acrshort{nocs} fails to localise two transparent objects (objs.~7 and 12). SegDD shows large inaccuracies for obj.~12 (margarita glass), obj.~7 (beer cup), and obj.~20 (deformed bottle), while \acrshort{lodeir} fails for objs.~20 and 15 (translucent bottle).
LoDE is less accurate with non-symmetric objects (\eg~obj.~20) and under challenging lighting (last three columns), but successfully estimates transparent objects (\eg~obj.12, margarita glass).

\section{Conclusion}
\label{sec:concl}

We proposed LoDE, a method to estimate the dimensions of container-like objects with circular symmetric shape, without relying on depth information, markers, or 3D models. LoDE uses an iterative multi-view 3D-2D shape fitting algorithm of a generative 3D sampling model, verifying the model on the object image masks of two wide-baseline cameras. 
To better handle transparent objects, LoDE uses a DNN-based semantic segmentation approach re-trained on selected high-level object classes of containers. For the evaluation, we collected a dataset of containers with different degrees of transparency, and under varying lighting conditions and backgrounds. The object localisation success ratio of LoDE is 86.96\% and its average error in estimating the objetc dimensions is smaller than 2~cm. 
As future work, we will generalise the approach to handle occlusions and generic object shapes under different poses. 

\bibliographystyle{IEEEbib}
\bibliography{main}

\end{document}